\documentclass[letterpaper]{article} 
\usepackage{aaai25}  
\usepackage{times}  
\usepackage{helvet}  
\usepackage{courier}  
\usepackage[hyphens]{url}  
\usepackage{graphicx} 
\usepackage{natbib}  
\usepackage{caption} 
\usepackage{algorithm}
\usepackage{algorithmic}
\usepackage{mathtools}
\usepackage{multirow}
\usepackage{amssymb}
\usepackage{enumitem}

\usepackage{newfloat}
\usepackage{listings}
\DeclareCaptionStyle{ruled}{labelfont=normalfont,labelsep=colon,strut=off} 
\floatstyle{ruled}
\newfloat{listing}{tb}{lst}{}
\floatname{listing}{Listing}
\title{Universal Features Guided Zero-Shot Category-Level Object Pose Estimation}
\author{
    Wentian Qu\textsuperscript{\rm 1,\rm 2,\rm 3},
    Chenyu Meng\textsuperscript{\rm 1,\rm 2},
    Heng Li\textsuperscript{\rm 3},
    Jian Cheng\textsuperscript{\rm 1,\rm 2},
    Cuixia Ma\textsuperscript{\rm 1,\rm 2},\\
    Hongan Wang\textsuperscript{\rm 1,\rm 2},
    Xiao Zhou\textsuperscript{\rm 4},
    Xiaoming Deng\textsuperscript{\rm 1,\rm 2}\thanks{indicates corresponding author.},
    Ping Tan\textsuperscript{\rm 3}\footnotemark[1]
}
\affiliations{
   \textsuperscript{\rm 1}Institute of Software, Chinese Academy of Sciences,
   \textsuperscript{\rm 2}University of Chinese Academy of Sciences,\\
   \textsuperscript{\rm 3}Hong Kong University of Science and Technology,
   \textsuperscript{\rm 4}Aerospace Information Research Institute, Chinese Academy of Sciences\\
    \{wentian2019, chengjian, cuixia, hongan, xiaoming\}@iscas.ac.cn, 
    mengchenyu21@mails.ucas.ac.cn,\\
    lh.heng.li@connect.ust.hk,
    zhouxiao@aircas.ac.cn, 
    pingtan@ust.hk
    
}

\usepackage{bibentry}

\begin{document}
\maketitle

\begin{figure*}[t]
\centering%
\includegraphics[width=.95\textwidth]{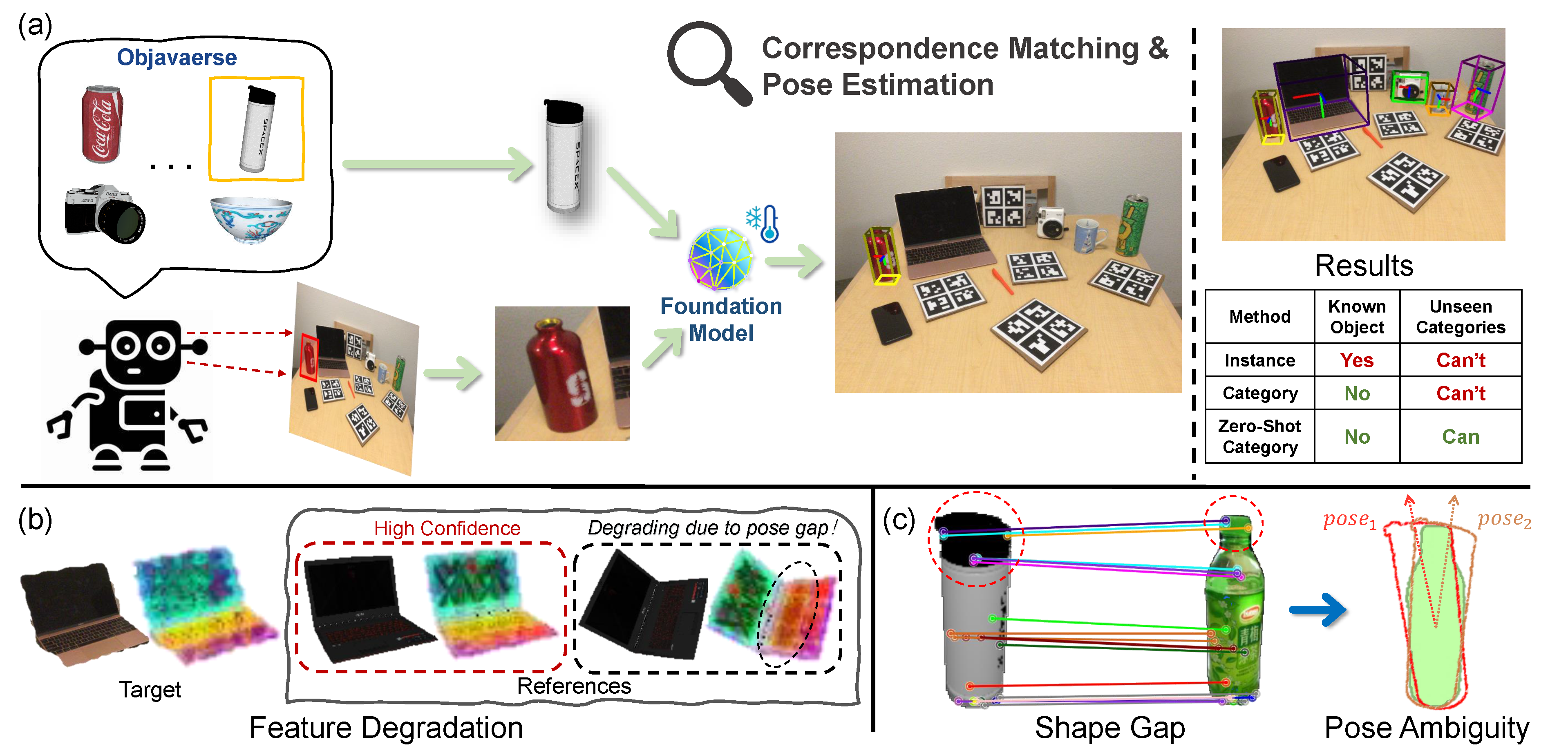}
\captionof{figure}{(a) We propose a zero-shot pose estimation method for unseen categories using universal features and obtain accurate results for multi-category scenes. Our method offers cost-efficient and superior generalization ability over traditional instance-level and category-level methods. (b) The correspondence with universal features degrades when pose has large gaps. (c) The shape gap between objects will cause pose ambiguity in optimization. These challenges affect the accuracy of pose estimation.}
\label{fig:teaser}
\end{figure*}
\begin{abstract}
Object pose estimation, crucial in computer vision and robotics applications, faces challenges with the diversity of unseen categories. We propose a zero-shot method to achieve category-level 6-DOF object pose estimation, which exploits both 2D and 3D universal features of input RGB-D image to establish semantic similarity-based correspondences and can be extended to unseen categories without additional model fine-tuning. Our method begins with combining efficient 2D universal features to find sparse correspondences between intra-category objects and gets initial coarse pose. To handle the correspondence degradation of 2D universal features if the pose deviates much from the target pose, we use an iterative strategy to optimize the pose. Subsequently, to resolve pose ambiguities due to shape differences between intra-category objects, the coarse pose is refined by optimizing with dense alignment constraint of 3D universal features. Our method outperforms previous methods on the REAL275 and Wild6D benchmarks for unseen categories. 
\end{abstract}
\begin{links}
    \link{Project Page}{https://iscas3dv.github.io/universal6dpose/}
\end{links}
\section{1. Introduction}
\label{sec:intro}

Object pose estimation, which aims to estimate the orientation and location of an object in 3D space, is a long-standing challenge and plays a key role in AR/VR and robotics applications. Instance-level methods~\cite{cosypose,posecnn,li2018deepim} achieve pose estimation for seen instances. 
To deal with unseen instances of the same category, category-level methods~\cite{zhang2022self,normalized,he2022towards} introduce a mean shape for a category object with improved model attention to texture and geometry details,
while these methods require retraining or model fine-tuning when applied to unseen categories.
To address this limitation, we aim to address zero-shot category-level object pose estimation from input RGB-D, which utilizes both 2D and 3D universal features from pre-trained models to achieve pose estimation for unseen categories.

Witnessed by a strong ability to establish correlations ~\cite{amir2021deep,zhang2022self} with 2D universal features extracted by the foundation models~\cite{dinov1,dinov2,stable_diffusion} pre-trained on large-scale datasets, recent work has explored 2D universal features to estimate object pose, such as FoundPose~\cite{ornek2023foundpose}, FoundationPose~\cite{wen2024foundationpose} and Zero-Pose~\cite{goodwin2022zero}.
FoundPose needs to know the model of the instance object in advance, and FoundationPose needs to use multi-view images to model the object before pose estimation. These settings require strong object shape priors and are difficult to extend to unseen categories.
Zero-Pose establishes 2D correspondences for object pose estimation using DINOv1~\cite{dinov1}. We find that DINOv1 is less effective in building correspondences and is severely affected by object pose (Fig.~\ref{fig:teaser} (b)). The correspondence will degrade if the pose difference between the reference image and the target image is large (Fig.~\ref{fig:iteration_effect}).
They also do not consider shape differences between intra-category objects, resulting in biased correspondences and inaccurate pose estimation (Fig.~\ref{fig:teaser} (c)).
Similarly to 2D universal features, 3D universal features, extracted by pre-trained models such as DGCNN~\cite{dgcnn} can provide effective geometric clues for correspondence, while they are barely used in object pose estimation. 
Inspired by this, we utilize both 2D and 3D universal features to solve the 6-DOF pose estimation for unseen categories without model training or fine-tuning, achieving a zero-shot category-level pose estimation. In particular, we do not need to know the 3D model of the instance objects. Our method offers superior generalizability over traditional instance-level and category-level methods~(Fig.~\ref{fig:teaser} (a)).

In this paper, we design a coarse-to-fine framework for accurate 6-DOF pose estimation. At the coarse stage, it identifies sparse correspondences to solve an initial coarse object pose. 
Given an input RGB-D image, we use a reference model of the interested category to render reference images and extract 2D universal features from both the target and rendered reference images. 
We then calculate the cosine similarity map between the 2D features and use cyclical distance to select Top-k correspondences. Combined with the depth map and camera intrinsics, we choose the Top-k keypoints in the camera coordinate and calculate the transformation from the reference to the target space to get the initial coarse 6-DOF object pose by a least-squares solution. 
To deal with the problem of feature correspondence degradation of 2D universal features if the initial pose deviates much from the target pose, we use an iterative strategy to optimize the correspondence and coarse pose.
After the coarse pose estimation, we map the reference model to the target image space to perform pose refinement with pixel-wise optimization. 
In order to resolve pose ambiguities due to shape differences between intra-category objects during the optimization, we employ 3D universal features extracted from the point cloud to refine the 6-DOF object pose and the reference model iteratively by dense pixel-level registration.

The main contribution of our method can be summarized as follows: 1) We propose a 2D/3D universal features guided zero-shot category-level object pose estimation with coarse-to-fine optimization. To deal with the correspondence degrade issue of 2D universal features, we use an iterative strategy to optimize the correspondence and coarse pose; 2) During the pose refinement, to handle pose ambiguity due to intra-category shape difference, we employ 3D universal features to refine the 6-DOF object and the shape of reference model by dense pixel-level registration; 3) Experiments on the REAL275~\cite{wang2019normalized} and Wild6D~\cite{ze2022category} benchmarks demonstrate that our method establishes robust correspondences based on pretrained 2D/3D universal features, resulting in accurate pose estimation based on coarse-to-fine optimization.

\section{2. Related Work}
\noindent \textbf{Instance-Level Object Pose Estimation.}
Instance-level object pose estimation methods regard each object as an independent entity with known object shapes. They directly regress the object pose within each RoI through characteristics~\cite{cosypose,posecnn,li2018deepim}, or estimate the object pose through 2D-3D correspondences based on conventional PnP~\cite{hodan2020epos,peng2019pvnet,tekin2018real} or PnP with network learning~\cite{hu2020single, wang2021gdr}. 
However, a major challenge for instance-level methods is that they struggle to estimate poses on unseen objects. 

\noindent \textbf{Category-Level Object Pose Estimation.}
Category-level object pose estimation methods divide objects into different categories~\cite{wang2022phocal,jung2024housecat6d}, emphasizing the commonality between objects. They perform well in 6-DoF pose estimation for unknown object shape~\cite{chen2024secondpose,gpv,lin2024instance}, and can directly learn the pose distribution of objects using the shape prior~\cite{
burchfiel2019probabilistic,sahin2018category}.
Previous methods either use a Normalized Object Coordinate Space (NOCS) representation \cite{normalized} to estimate object pose~\cite{chen2020learning,ze2022category,fsnet}, or learn the pose distribution using geometric priors such as shape~\cite{
tian2020shape}, symmetry~\cite{lin2021donet} and keypoints~\cite{lin2022single}.
The existing category-level methods suffer from model generalization to unseen categories.
\begin{figure*}[ht]
\centering%
\includegraphics[width=\linewidth]{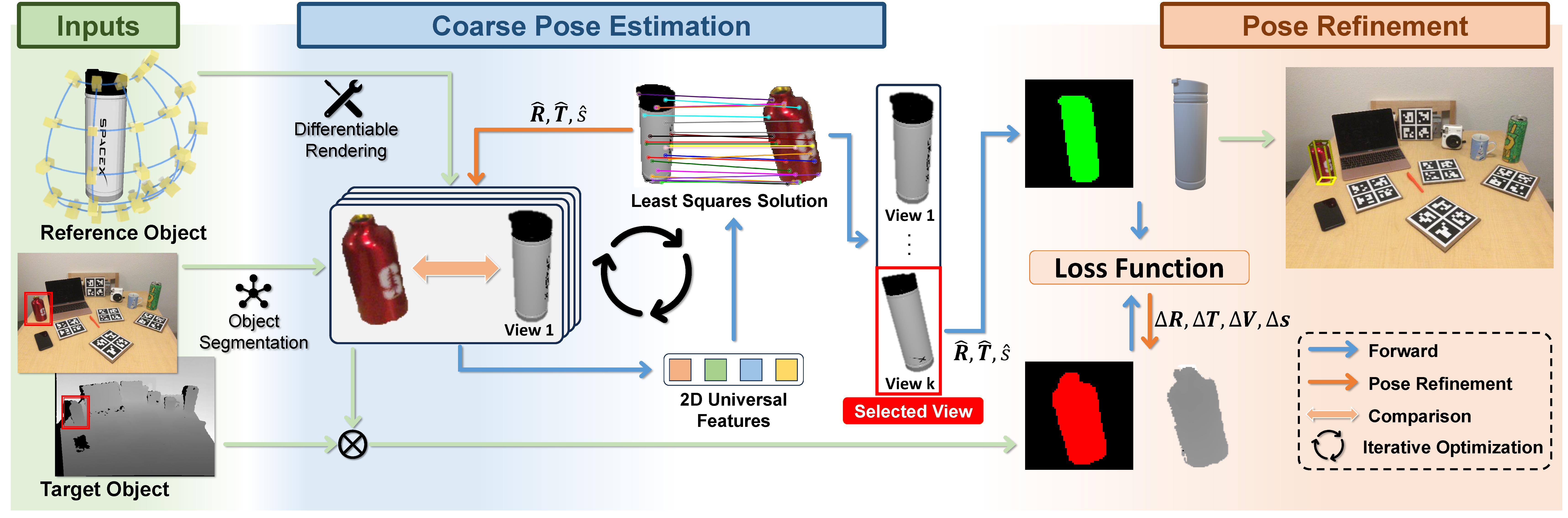}
\caption{Overview. Our framework includes a keypoint-level coarse pose estimation module and a pixel-level pose refinement module. In the first module, we establish the correspondences between image pairs based on the 2D universal features and calculate the coarse pose using least squares in an iterative manner. In the second module, we use pixel-level optimization combined with 3D universal features to refine the pose and shape of reference model to obtain the fine pose.
}
\label{fig:pipeline}
\end{figure*}

\noindent \textbf{Zero-Shot Object Pose Estimation.}
In order to remove the time-consuming dataset collection requirement, several works leverage the foundation models such DINO~\cite{dinov1,dinov2} and Stable Diffusion (SD) ~\cite{stable_diffusion} for pose estimation~\cite{goodwin2022zero,chen2023zeropose}.    
Zero-shot instance-level methods ~\cite{ornek2023foundpose,fan2023pope,labbe2023megapose} estimate the object pose of the seen instance through feature matching based on 2D universal features such as DINO. However, these methods cannot be applied to unknown instances.   
In order to overcome the generalization limitation of these instance-level methods on known object instances, the zero-shot category-level method~\cite{goodwin2022zero} is designed to be generalized to unseen objects based on universal features DINOv1. However, this method is affected by less effective features, especially in varied pose and shape, resulting in inaccurate pose results, and it is verified that DINOv1 is inferior to more advanced universal features such as DINOv2 and the combination with Stable Diffusion features~\cite{zhang2023tale,luo2023diffusion,tang2023emergent}.
Our method has two differences from \cite{goodwin2022zero}. 
First, to deal with the inefficient of DINOv1 in which the feature similarity performance drops under a large pose gap, we conduct a strong universal feature combined with DINOv2 and SD, and design an iterative module to conduct feature matching under step-by-step optimized pose.
Second, to address the shape gaps, we design a pixel-level pose refinement framework, which utilizes 3D universal features to jointly optimize the object's pose and shape with a new universal alignment constraint.

\section{3. Method}
This paper exploits multi-modal (both 2D and 3D) universal features to estimate object pose on unseen categories. Our method strives to address the following three challenges,
1) to overcome the correspondence degradation of 2D universal features caused by large pose differences;
2) to exploit the effect of 3D universal features for category-level object pose estimation;
3) to handle the pose ambiguity caused by shape differences between the reference and target object. 

Our zero-shot category-level pose estimation works with multi-modal universal features~(Sec.~3.1) in a coarse-to-fine manner (See Fig.~\ref{fig:pipeline}), which first utilizes 2D universal features for keypoint-level coarse pose estimation~(Sec~3.2) and then uses 3D universal features for pixel-level pose refinement~(Sec.~3.3).  
To solve the first challenge, we propose an iterative optimization to establish more accurate correspondences based on the rendered reference image under updated pose~(Sec~3.2).
To solve the last two issues, we perform a pixel-level pose refinement based on 3D universal features with a new universal alignment constraint to resolve the ambiguity between shape and pose optimization~(Sec~3.3). 

\subsection{3.1 Multi-Modal Universal Features}
\noindent \textbf{2D Universal Features.}
The 2D universal features are extracted by pre-trained image models such as DINOv1~\cite{dinov1}, DINOv2~\cite{dinov2}, and Stable Diffusion (SD)~\cite{stable_diffusion}.
They provide effective texture prior information to establish semantic correspondence between images. 
Given a target image $\mathbf{I}_{t} \in \mathbb{R}^{H \times W \times 3}$, we use Mask R-CNN~\cite{maskrcnn} to predict the shape token of the interested object and retrieve a reference mesh model from Objaverse~\cite{deitke2023objaverse} to render the reference image.
Given the image, we extract the universal 2D features $\mathbf{F}$ by sending them to DINOv1, DINOv2 and SD:
\begin{equation}
  \mathbf{F} = (\alpha_{D1} \Vert \mathbf{F}_{D1} \Vert_{2},\alpha_{D2} \Vert \mathbf{F}_{D2} \Vert_{2}, \alpha_{SD} \Vert \mathbf{F}_{SD} \Vert_{2})
\label{eq:cat_feature}
\end{equation}
where $\mathbf{F}_{D1}$, $\mathbf{F}_{D2}$, $\mathbf{F}_{SD}$ denotes the features from DINOv1, DINOv2 and SD, and $\alpha$ denotes the hyperparameter that balances the proportions of different 2D universal features. 
We conduct a systematic study on pose estimation over different combinations of 2D universal features and observe that the combination of DINOv2 and SD can leverage the strengths of both local and global semantic similarity to generate robust correspondences, which result in an accurate pose.

\noindent \textbf{3D Universal Features.}
DGCNN~\cite{dgcnn} pre-trained on 3D point cloud datasets could extract 3D universal features $\mathbf{F}_{3d}$ containing geometric information. 
As shown in the red dashed box of Fig.~\ref{fig:pose_refine}(a), the 3D universal features aligned based on the initial pose contain more semantic similarity in geometric details, which can further refine object pose and resolve the shape ambiguities.

\subsection{3.2 Keypoint-Level Coarse Pose Estimation}
\label{sec:method_2}
We extract the 2D features for reference and target images and leverage them to estimate the coarse pose transformation by computing the sparse keypoint-level correspondence.

\noindent \textbf{Establish Correspondence.} 
Specifically, we render four images $\{\mathbf{I}_{r}\}$ of the reference object from the front, back, and two sides. Then, we extract 2D feature for all the reference images $\{\mathbf{I}_{r}\}$ and the target image $\mathbf{I}_{t}$, denoted by $\{\mathbf{F}(\mathbf{I}_{r})\}$ and 
$\mathbf{F}(\mathbf{I}_{t})$, respectively. A score matrix $\mathbf{S}$ can be defined based on the cosine similarity between the features:
\begin{equation}
        \mathbf{S}(p,q) = d_{cos}(\mathbf{F}(\mathbf{I}_{t})_{p},\mathbf{F}(\mathbf{I}_{r})_{q}), p \in [1,\mathbf{N}_t], q \in [1, \mathbf{N}_{r}],
\end{equation}
where $\mathbf{N}_{t}$ and $\mathbf{N}_{r}$ are the number of patches in the target and the reference feature maps, $d_{cos}(\cdot,\cdot)$ is the cosine similarity between two vectors.
The cyclical distance matrix $\mathbf{D}\in \mathbb{R}^{N_{t}}$ based on $\mathbf{S}$ can be used to compute the correspondence for each patch pair between the reference and target features:
\begin{equation}
    \mathbf{D}_{p,p\in[1:\mathbf{N}_{t}]} = d(p, \mathop{\arg\max}_{p'\in [1:N_{t}]}\mathbf{S}(p',\mathop{\arg\max}_{q\in[1:N_{r}]}\mathbf{S}(p,q))),
\end{equation}
where $d(\cdot,\cdot)$ is the L2 distance. We select $M$ correspondence based on ascending order of $\mathbf{D}$ for each image pair. 

\begin{figure}[t]
\centering%
\includegraphics[width=\linewidth]{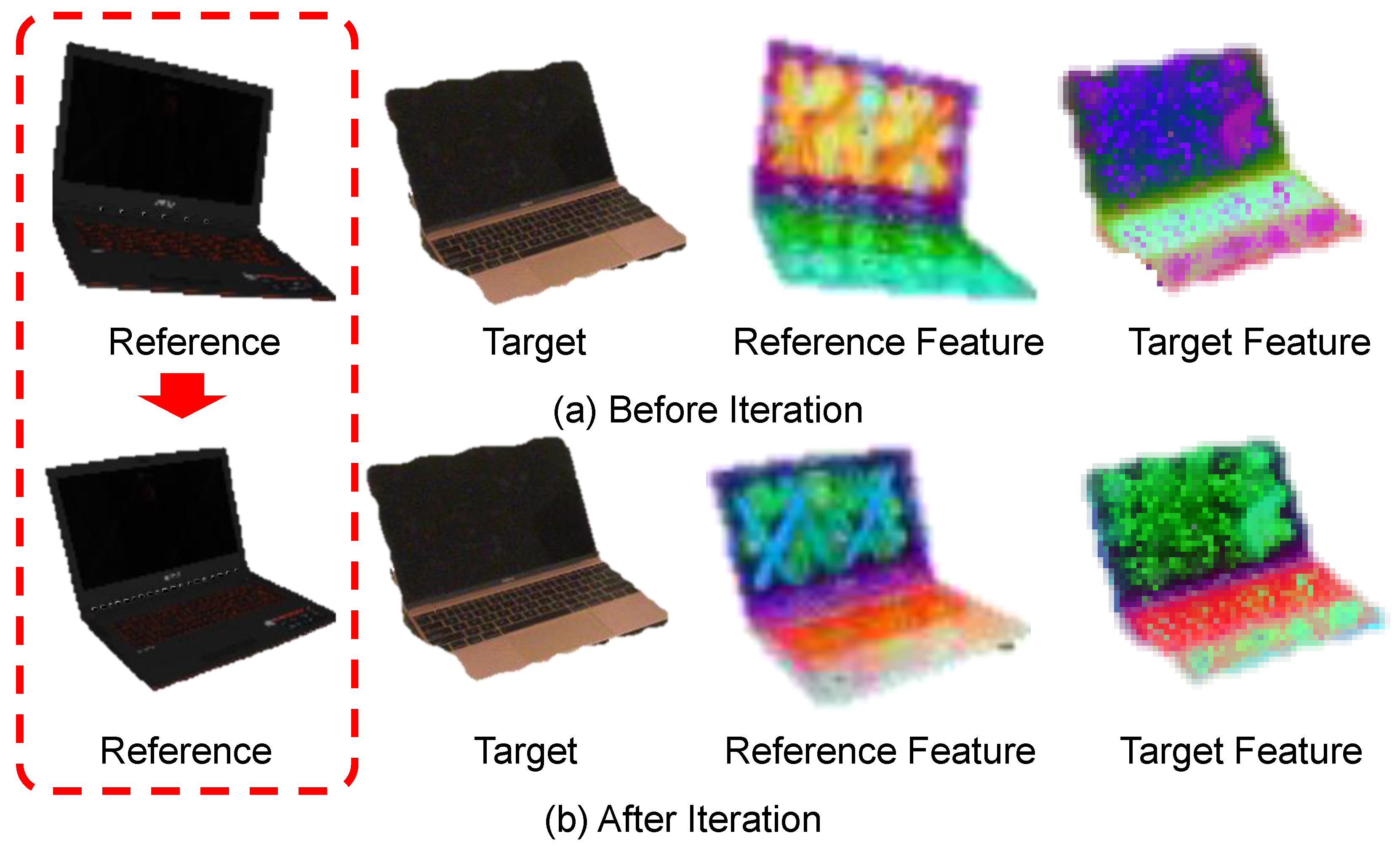}
\caption{Feature Performance Drop and Effect of Iterative Estimation. When there are large pose differences between objects, the 2D universal features similarity degrade. After iterative optimization, as the objects are gradually aligned, the correspondence between the objects become smoother, which support to calculate an accurate pose.}
\label{fig:iteration_effect}
\end{figure}
\noindent \textbf{Iterative Coarse Pose Estimation.} 
For each reference and target image pair, we can lift the 2D keypoints into 3D camera space to get the reference point cloud $\mathbf{P}_{r} \in \mathbb{R}^{N \times 3}$ and target 3D point cloud $\mathbf{P}_{t} \in \mathbb{R}^{N \times 3}$ using the depth map and camera intrinsics. Then we use Umeyama~\cite{umeyama1991least} with RANSAC~\cite{fischler1981random} to compute the updated object pose for each reference image:
\begin{equation}
  (\hat{\mathbf{R}},\hat{\mathbf{T}},\hat{s}) = \mathop{\arg\min}\limits_{\mathbf{R},\mathbf{T},s}\frac{1}{2}\sum^{M}_{m=1}\Vert\mathbf{P}^{m}_{t}-(s\mathbf{R}\mathbf{P}^{m}_{r}+\mathbf{T})\Vert^{2}_{2},
\label{eq:ransac}
\end{equation}
where $\hat{\mathbf{R}} \in \mathbb{R}^{3 \times 3}$,$\hat{\mathbf{T}} \in \mathbb{R}^{3}$, $\hat{s}\in \mathbb{R}$ represent the object rotation, translation, and scale, respectively. 
We find that the semantic similarity based on 2D universal features will degrade when the pose between reference and target objects has large differences (Fig.~\ref{fig:iteration_effect} (a)). 
To solve this problem, we render the reference model with the updated coarse pose and establish more accurate correspondences to iteratively optimize the coarse pose. During the iteration, as the objects are gradually aligned, the correspondence between the objects become more consistent, which supports the calculation of an accurate pose~(Fig.~\ref{fig:iteration_effect} (b)). 
We define the confidence for each reference image by averaging the cosine similarity of the $M$ correspondences, and we choose the result with the highest confidence as the final coarse pose output.

\subsection{3.3 Pixel-Level Pose Refinement}
\label{sec:method_3}
\begin{figure}[ht]
\centering%
\includegraphics[width=\linewidth]{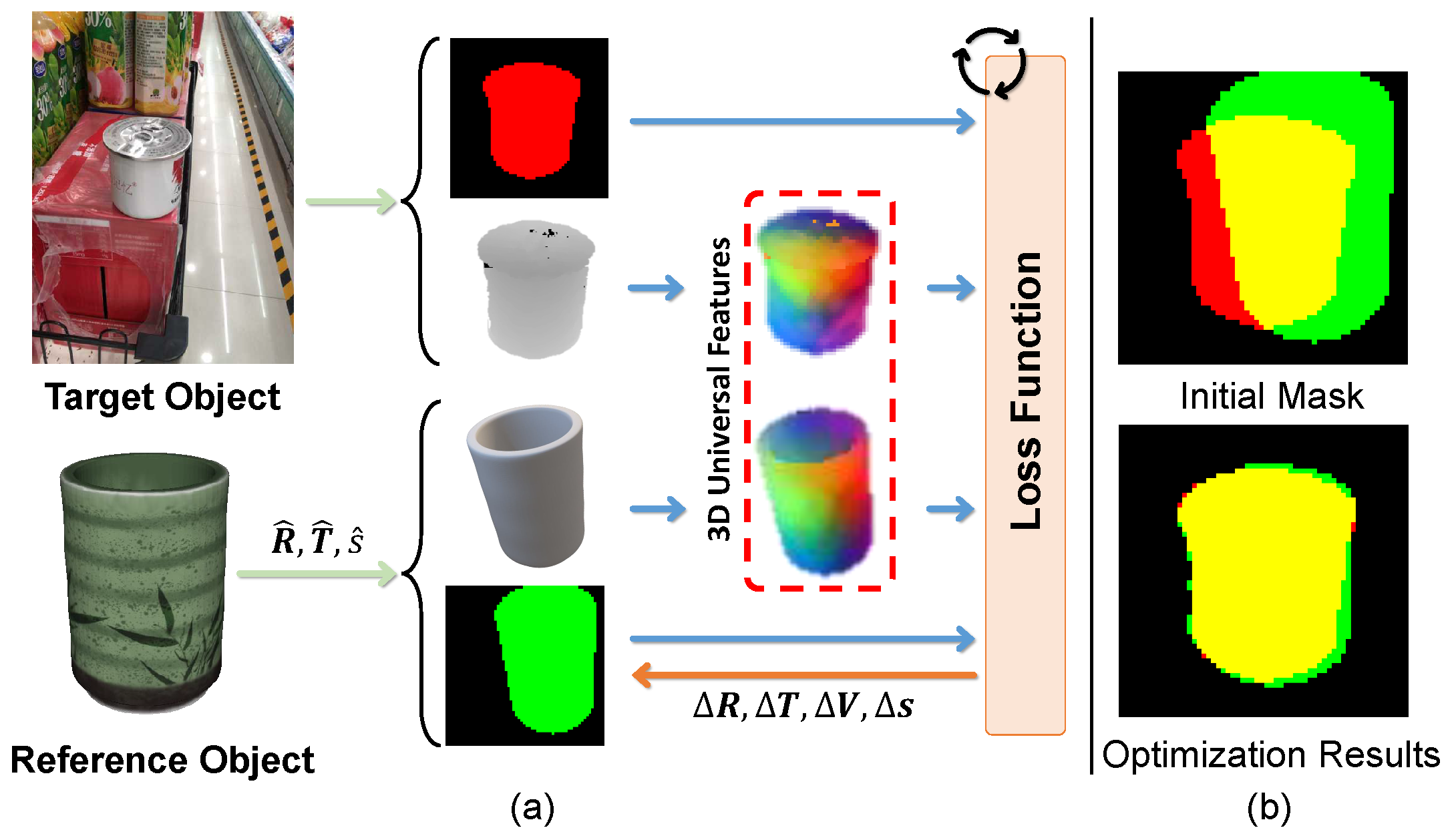}
\caption{(a) Pose Refinement. Based on the coarse pose as initialization, the reference model can be warped to the target space to obtain the initial mask and extract 3D universal features. Then we optimize the coarse pose and shape by minimizing the loss function. (b) After pose refinement stage, the pose and shape of the reference model are more accurately aligned with the target object.}
\label{fig:pose_refine}
\end{figure}

Although the keypoint-level method can achieve good results, estimating accurate object pose with a standard reference shape model for intra-category objects is still challenging, especially when the intra-category shape gap is large. Moreover, the keypoint-level method only uses sparse keypoints, thus it does not utilize dense geometric information to reduce the pose searching space. To address these two issues, we propose a pose refinement module by jointly optimizing object shape and pose (Fig.~\ref{fig:pose_refine} (a)), a dense pixel-level optimization built on 3D universal features. After optimizing shape and pose, the objects are more accurately aligned~(Fig.~\ref{fig:pose_refine} (b)).

\noindent \textbf{3D Universal Features Extraction.} 
We use the coarse pose to align the reference object with the target object, and then feed them to the pre-trained model DGCNN to extract 3D universal features, respectively. Compared with 2D features, the 3D universal features can measure the 3D geometric semantic similarity for intra-category objects, which tackle the searching ambiguity in pose and shape.

\noindent \textbf{Optimization Parameters.}
During pose refinement for each instance, we aim to optimize the rotation $\Delta \mathbf{R} \in \mathbb{R}^{3 \times 3}$, translation $\Delta \mathbf{T} \in \mathbb{R}^{3}$, independent deformation $\Delta \mathbf{V} \in \mathbb{R}^{N \times 3}$ for each vertex and scale $\Delta \mathbf{s} \in \mathbb{R}^{3}$.
The optimized object rotation and translation can be expressed as $\Bar{\mathbf{R}} = \Delta \mathbf{R} \times \hat{\mathbf{R}}, \Bar{\mathbf{T}} = \Delta \mathbf{T} + \hat{\mathbf{T}}$, where we abbreviate $\hat{s} \cdot \hat{\mathbf{R}}$ as $\hat{\mathbf{R}}$ in Eq.~\ref{eq:ransac}. We define the reference mesh vertices in model coordinate as $\mathbf{V}$ and the optimized vertices can be formulated as $\Bar{\mathbf{V}}=e^{\Delta \mathbf{s}}(\mathbf{V}+\Delta \mathbf{V})$. 

\noindent \textbf{Loss Function.} 
We define the loss to constrain the output pose and object shape to guarantee feasible results, including pose optimization loss $L_{p}$ and regularization loss $L_{r}$. The purpose of pose optimization loss is to optimize the pose and shape of the reference object to align accurately with the target object, and regularization loss is used to constrain the optimized reference object to be close to the initial state. The total loss function can be defined as: $L = L_{p} + L_{r}$.

\noindent \textbf{Pose Optimization Loss $L_{p}$.} 
We solve the relative motion between the target and reference objects by minimizing a registration objective function defined to match projected masks, 3D universal features and shapes. Our pose optimization loss including mask loss $L_{m}$, Chamfer loss $L_{c}$ and universal alignment loss $L_{g}$ as: $L_{p} = \alpha_{m}L_{m} + \alpha_{c}L_{c} + \alpha_{g}L_{g}$.

The mask loss measures the difference between the reference mask $\mathbf{M}_{r}$  and the target mask $\mathbf{M}_{t}$, which can be calculated as: $L_{m} = 1 - \frac{\mathbf{M}_{r}\cap\mathbf{M}_{t}}{\mathbf{M}_{r}\cup\mathbf{M}_{t}}$. 
We also use Chamfer loss $L_{c}$ to enforce that the position and shape difference from the reference point cloud to the target point cloud is small.
The 3D universal features reflect the relative positional relationship between the sampling points and the global geometry, which contain more shape details.
Therefore, we propose a new universal alignment loss $L_{g}$ to ensure that the 3D universal features between the reference object and the target object are consistent, which can solve the ambiguity when optimizing the pose and shape of the reference model.
We calculate the cosine similarity of 3D universal features of the reference model and the target point cloud, and then make the 3D positions between high-confidence keypoint pairs as close as possible. The universal alignment loss can be defined as: $L_{g} = \frac{1}{2}\sum_{d_{cos}^{3D}(p^{r},p^{t})>\beta_{g}}\Vert p^{r}-p^{t} \Vert^{2}_{2}$, where $p^{r}$ and $p^{t}$ represent the reference and target object point cloud, respectively, $d_{cos}^{3D}(p,q)$ represents the cosine similarity of 3D universal features between $p^{r}$ and $p^{t}$. The hyperparameter $\beta_{g}$ controls the confidence threshold and we set it to 0.8.

\noindent \textbf{Regularization Loss $L_{r}$.} 
During the optimization, we also want to constrain the optimized reference model not to deviate greatly from the initial state.
We use pose regularization loss to enforce the refined pose to be close to the initial pose and use center point regularization loss to reduce the object displacement before and after optimization. The deformation regularization loss is used to constrain the deformation of the vertices to be small. In addition, we follow Pytorch3D~\cite{ravi2020pytorch3d} to minimize geometric distortion with normal, edge, and Laplacian constraints.

\begin{figure*}[h]
\centering%
\includegraphics[width=\linewidth]{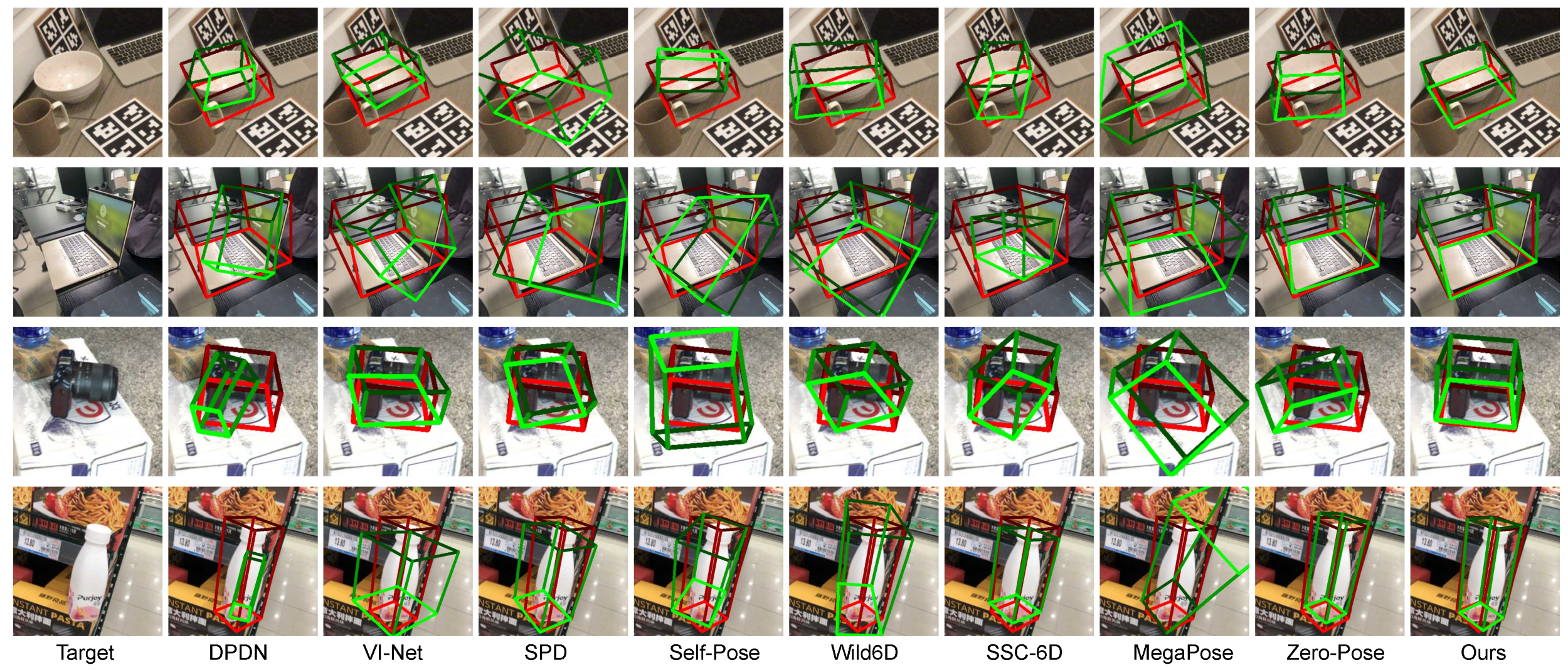}
\caption{Qualitative results on REAL275 and Wild6D. The red box represents the ground truth, and the green box represents the estimation. Previous methods exhibit large errors when applied to unseen categories due to the significant texture and shape differences.
Our method demonstrates strong generalization on unseen categories with accurate pose estimation.}
\label{fig:qual_res}
\end{figure*}
\begin{table*}[h]
\centering
\fontsize{9}{11}\selectfont
\begin{tabular}{c|c|c|c|c|c|c|c}
\hline
\multirow{2}{*}{Method} & \multicolumn {3}{c|}{REAL275} & \multicolumn {3}{c|}{WILD6D} & \multicolumn {1}{c}{ }\\
\cline{2-8}
{}& $IOU_{0.25/0.5}$ & $5^{\circ}2cm$ / $5^{\circ}5cm$ & $10^{\circ}2cm$ / $10^{\circ}5cm$& $IOU_{0.25/0.5}$ & $5^{\circ}2cm$ / $5^{\circ}5cm$ & $10^{\circ}2cm$ / $10^{\circ}5cm$ & Speed\\
\hline
DPDN &{70.35/12.36}&{5.11}/{8.27}&{12.86}/{20.32}&{74.73/15.77}&{6.82}/{11.86}&{15.58}/{27.44}&{0.01}\\
VI-Net &{51.79/13.81}&{19.47}/\textbf{37.52}&{22.59}/{50.79}&{60.59/25.40}&{42.14}/\textbf{66.15}&{44.44}/{72.84}&{0.01}\\
SPD &{45.91/8.81}&{4.18}/{6.49}&{10.33}/{16.88}&{37.95/8.55
}&{1.99}/{6.10}&{5.76}/{16.35}&{0.01}\\
Self-Pose &{60.54/ 6.96}&{0.40}/{0.48}&{1.44}/{1.98}&{63.34/ 3.61}&{0.10}/{0.98}&{0.38}/{3.44}&{0.01}\\
Wild6D &{48.84/12.89}&{5.26}/{7.71}&{12.28}/{18.64}&{55.50/15.36}&{10.10}/{16.42}&{20.23}/{36.32}&{0.01}\\
SSC-6D &{59.84/11.21}&{4.73}/{6.54}&{12.41}/{17.36}&{69.81/15.82}&{6.30}/{10.29}&{16.33}/{27.70}&{0.08}\\
MegaPose &{5.93/0.27}&{0.28}/{0.93}&{0.45}/{2.16}&{0/0}&{0}/{0}&{0}/{0.01}&{0.92}\\
Zero-Pose & \textbf{82.51}/{58.34}& {21.36}/{22.94}&{43.84}/{49.82}& 
{86.30/55.36}& {24.89}/{43.56}&{41.38}/{71.09}&{0.97}\\
Ours&{80.06}/\textbf{63.49}&\textbf{30.61}/{33.23}&\textbf{50.15}/\textbf{57.74}&\textbf{88.46}/\textbf{67.16}&\textbf{47.69}/{60.58}&\textbf{59.46}/\textbf{80.47}&{3.83}\\

\hline
\end{tabular}
\caption{Quantitative results on REAL275 and Wild6D. Zero-Shot methods show strong generalization ability on unseen categories and our cascaded coarse-to-fine optimization strategy is more effective in accurate pose prediction}
\label{table:quan_res}
\end{table*}

\section{4. Experiment}
\subsection{4.1 Experimental Setup}
\label{sec:exp_1}
\noindent \textbf{Datasets.}
We select Wild6D~\cite{ze2022category} and REAL275~\cite{normalized} for category-level object pose estimation. Wild6D provides 5,166 videos across 1,722 different objects with five categories (including bottles, bowls, cameras, laptops and mugs). REAL275 contains six testing scenes with six categories(including bottles, bowls, cameras, cans, laptops and mugs). Both benchmarks provide RGB-D images, and the foreground segmentation and shape token are obtained by Mask R-CNN~\cite{maskrcnn}. 

\noindent \textbf{Evaluation Metrics.}
We follow REAL275~\cite{normalized} to use intersection over union (IoU) with a threshold of 25\% ($IOU_{0.25}$) and 50\% ($IOU_{0.5}$) to evaluate 3D object localization. For object pose estimation, we report the average pose accuracy of unseen categories where a pose is considered accurate when the translation errors are smaller than $m~cm$ and rotation errors are smaller than $n^{\circ}$. We also compare the inference speed ($s$) of one frame.

\noindent \textbf{Baselines.}
We compare our method with three types of baselines: 1) \emph{Supervised Methods} train the models with ground truth pose annotations, including DPDN~\cite{dpdn}, VI-Net~\cite{vinet}and SPD~\cite{tian2020shape}. 2) \emph{Self-Supervised Methods} train the models without ground truth pose annotations, including Self-Pose~\cite{zhang2022self}, Wild6D~\cite{ze2022category} and SSC-6D~\cite{SSC}. 3) \emph{Zero-Shot Methods} without additional training including Zero-Pose~\cite{goodwin2022zero} and MegaPose~\cite{labbe2023megapose}.

\noindent \textbf{Experimental Settings.}
During comparison with baselines, the pose estimation models are not trained on the tested categories.
We follow REAL275~\cite{normalized} and Wild6D~\cite{ze2022category} to divide the training set and the test set. We adapt the leave-1 strategy for supervised methods, which selects one category as the test set and use the remaining categories to train the model. We conduct leave-1 experiments for each category and finally take the average of them. We directly use the official pre-trained models to conduct the leave-p experiments for self-supervised methods, as they have the property of per-subject-per-train. We select the model for each category, test it on other unseen categories, and finally take the average of evaluation metrics. For the zero-shot method, we test directly on all categories without model training or fine-tuning. 
During testing, we do not provide supervised and self-supervised methods with a reference model used in the zero-shot method as the shape prior. Because these methods have not seen such category shapes during training, it is impossible to establish an effective association with the target object and will result in poor prediction results.

\subsection{4.2 Implementation Details}
\label{sec:exp_2}
We set $\alpha_{D1}, \alpha_{D2},\alpha_{SD}$ to 0, 0.7, 0.3 in~Eq.\ref{eq:cat_feature}, 
set the hyperparameters 
$\alpha_{m}$, $\alpha_{g}$, $\alpha_{c}$ to 1, 1 and 0.1 respectively.
We iteratively update the object pose 2 times to obtain the coarse object pose and run RANSAC up to 1,000 times for each iteration to handle outliers. In the pose refinement stage, we use Adam~\cite{adam} as the optimizer to minimize the loss function.
We test our method on a single GeForce RTX 4090, costing 11.7 GB memory in coarse pose estimation stage and costing 5.5 GB memory in pose refinement stage for each instance.
Please refer to the supplementary file for more details.

\subsection{4.3 Comparison Results}
\label{sec:exp_3}
We show comparison results on REAL275 and Wild6D in Tab.~\ref{table:quan_res} and Fig.~\ref{fig:qual_res}. 
Supervised and self-supervised methods over-fit the seen category shape priors, which make them difficult to accurately predict the exact geometry of unseen categories, leading to significant errors and pose drifting. These non-foundation models cannot establish effective semantic similarities on unseen categories with complex texture and shape variations. 
VI-Net overfits symmetric objects and obtains a better $5^{\circ}5cm$ result. However, when the tolerance for rotation error increases ($10^{\circ}5cm$), its accuracy becomes worse than ours. 
Note that MegaPose fails in this experimental setup. We analyze that this instance-level zero-shot method cannot establish effective associations between different instances and the difference in scale between the reference object model and the target object leads to a large translation error.
Our method consistently outperforms Zero-Pose in almost all evaluation metrics, because our universal features can find more consistent correspondences (last row of Tab.~\ref{table:feature_comb}), and the cascaded coarse-to-fine optimization strategy is more effective in accurate pose prediction (second and fifth row of Tab.~\ref{table:dif_module}).  

\subsection{4.4 Ablation Results}
\label{sec:exp_4}
\begin{table*}[h]
\centering
\fontsize{9}{11}\selectfont
\begin{tabular}{c|c|c|c|c|c|c|c|c|c}
\hline
\multirow{2}{*}{Ite.} & \multirow{2}{*}{Ref.} & \multirow{2}{*}{$L_{g}$}& \multirow{2}{*}{Def.} & \multicolumn {3}{c|}{REAL275} & \multicolumn {3}{c}{WILD6D}\\
\cline{5-10}
{}&{}&{}&{}& $IOU_{0.25/0.5}$ & $5^{\circ}~2/5cm$& $10^{\circ}~2/5cm$& $IOU_{0.25/0.5}$ & $5^{\circ}~2/5cm$& $10^{\circ}~2/5cm$\\
\hline

{1}&{$\times$}&{$\times$}&{$\times$}
&{77.49}/{59.22}&{24.16}/{27.10}&{47.43}/{54.39}
&{87.17}/{60.36}&{32.79}/{52.63}&{49.08}/{79.55}\\
{2}&{$\times$}&{$\times$}&{$\times$}
&{80.00}/{58.39}&{28.56}/{30.94}&{48.33}/{55.27}
&{88.17}/{62.77}&{42.71}/{57.58}&{55.99}/\textbf{80.69}\\
{2}&{\checkmark}&{$\times$}&{\checkmark}
&{80.04}/{63.44}&{30.35}/{33.13}&{49.91}/{57.66}
&{88.45}/{67.11}&{47.48}/{60.37}&{59.35}/{80.42}\\
{2}&{\checkmark}&{\checkmark}&{$\times$}
&{80.04}/{58.94}&{26.74}/{29.18}&{44.05}/{54.18}
&{88.32}/{64.79}&{45.83}/{59.27}&{58.00}/{80.19}\\
{2}&{\checkmark}&{\checkmark}&{\checkmark}
&\textbf{80.06}/\textbf{63.49}&\textbf{30.61}/\textbf{33.23}&\textbf{50.15}/\textbf{57.74}
&\textbf{88.46}/\textbf{67.16}&\textbf{47.69}/\textbf{60.58}&\textbf{59.46}/{80.47}\\
\hline
\end{tabular}
\caption{Ablation studies on the influence of the iterative optimization and pose refinement module. After iteration, objects can be further aligned to establish smoother correspondences to get performance improvements. After pose refinement, the shapes between the objects are more similar, thereby ensuring more accurate object alignment to get precise pose estimation.}
\label{table:dif_module}
\end{table*}
\begin{table}[h]
    \centering
    \fontsize{9}{11}\selectfont
    \begin{tabular}{c|c|c|c|c}
    \hline
    & Method & $IOU_{0.25/0.5}$ & $5^{\circ}~2/5cm$ & $10^{\circ}~2/5cm$\\
    \hline
    
    \multirow{5}{*}{\rotatebox{90}{\begin{tabular}[c]{@{}c@{}}REAL275\end{tabular}}}&
    
    v1&{82.41/59.06}&{21.34}/{23.03}&{44.49}/{50.68}\\
    &v1+SD&{81.19/58.20}&{21.35}/{23.06}&{44.46}/{50.34}\\
    &v2&{76.47/56.96}&{21.71}/{24.95}&{45.04}/{53.00}\\
    &All&\textbf{84.62}/\textbf{62.77}&{23.36}/{25.02}&{47.02}/{53.77}\\
    &v2+SD&{77.49}/{59.22}&\textbf{24.16}/\textbf{27.10}&\textbf{47.43}/\textbf{54.39}\\
    \hline\hline
    
    \multirow{5}{*}{\rotatebox{90}{\begin{tabular}[c]{@{}c@{}}WILD6D\end{tabular}}}&
    
    v1&{85.36/56.20}&{23.81}/{42.18}&{41.68}/{70.85}\\
    &v1+SD&{85.53/56.30}&{23.68}/{42.77}&{41.61}/{70.77}\\
    &v2&{83.87/56.90}&{30.94}/{46.86}&{46.36}/{74.05}\\
    &ALL&{86.46/57.73}&{24.51}/{43.72}&{42.03}/{72.45}\\
    &v2+SD&\textbf{87.17}/\textbf{60.36}&\textbf{32.79}/\textbf{52.63}&\textbf{49.08}/\textbf{79.55}\\
    \hline
    
    \end{tabular}
    \caption{Ablation studies on different combinations of 2D universal features. We find that the feature combination of DINOv2 and SD will comprehensively utilize both global and local context to obtain more accurate estimation results.}
    \label{table:feature_comb}
    \end{table}

\noindent \textbf{Effect of Iterative Optimization.}
To investigate the effect of iterative optimization in coarse pose estimation stage, we evaluate the pose accuracy at different iterative steps.
As shown in Tab.~\ref{table:dif_module} (Iter.), by comparing the results in the first and second rows, we conclude that iterative optimization can significantly improve the performance of the object pose (i.e. $5^{\circ}2cm$ and $5^{\circ}5cm$). The iterative optimization can effectively address the problem of correspondence degradation if the pose deviates much from the target pose: As the estimated object pose improves, better correspondences will be established between the target object and the reference object, making the pose estimation more accurate~(Fig.~\ref{fig:iteration_effect}). 

\begin{figure}[ht]
\centering%
\includegraphics[width=\linewidth]{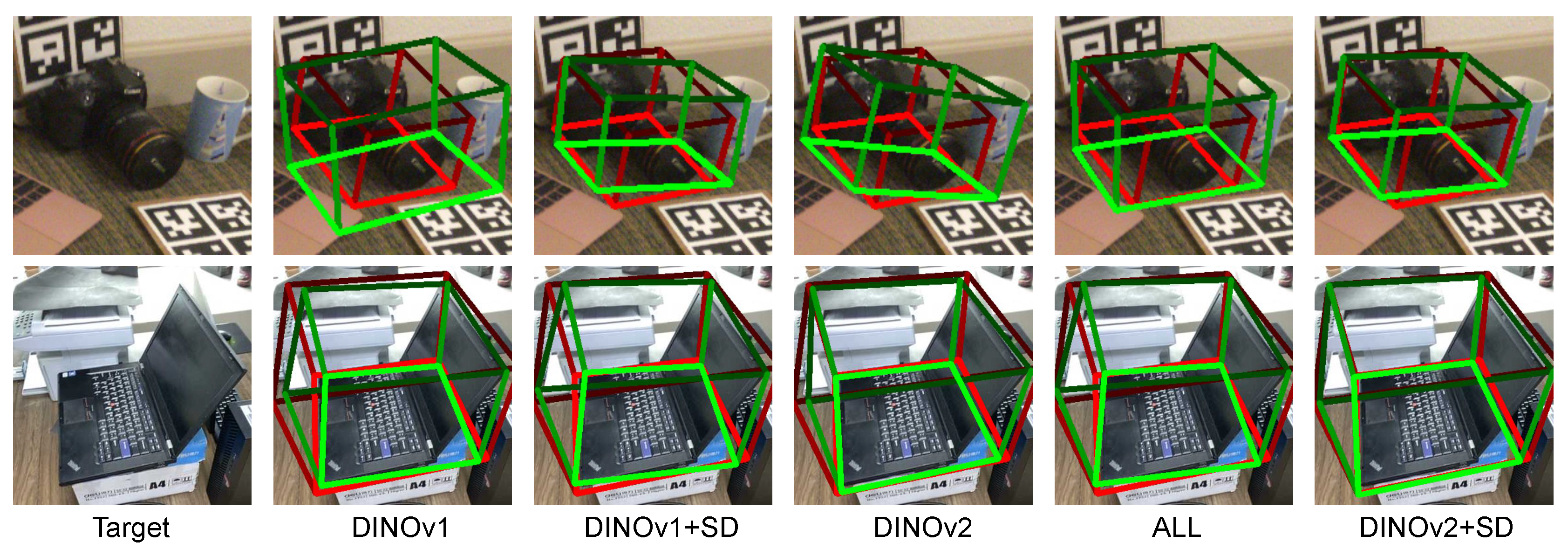}
\caption{Qualitative results on different combinations of 2D universal features. 
}
\label{fig:ablation}
\end{figure}
\begin{figure}[ht]
\centering%
\includegraphics[width=\linewidth]{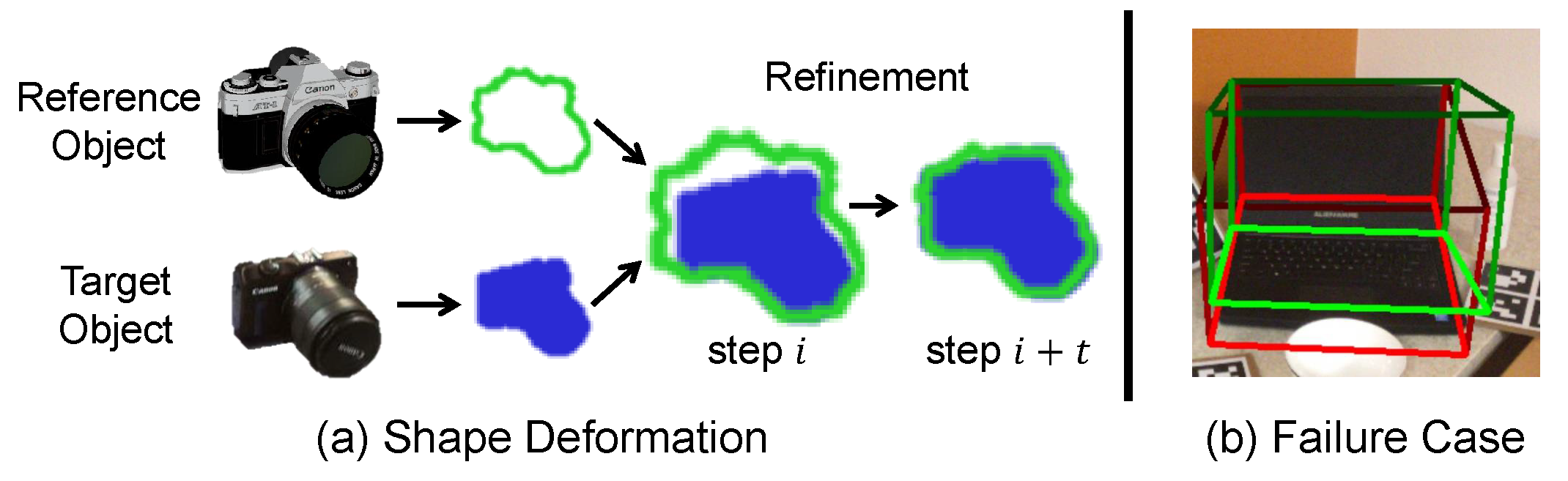}
\caption{(a) After the shape optimization, the reference object shape will become closer to the target object shape, resulting in more accurate pose. (b) The occlusion makes the shape of the target object incomplete, causing translation and rotation errors.}
\label{fig:shape_def}
\end{figure}

\noindent \textbf{Effect of Pose Refinement Module.}
We compare the results with and without the pose refinement module to investigate the effect of the pose refinement module. As shown in the last row of Tab.~\ref{table:dif_module}, the pose refinement can further improve the pose accuracy. 
A critical step in pose refinement is shape deformation. Fig.~\ref{fig:shape_def} (a) shows the reference shape before and after deformation. The camera has been optimized to low the top bulge, and shrunk the overall body.
After shape deformation, the bias in correspondences caused by shape differences is reduced, resulting in a better alignment and more accurate pose estimation.
By comparing the third row and the fifth row in Tab.~\ref{table:dif_module}, we observe that our universal alignment loss $L_{g}$ allows for more accurate pose. Conceptually, 3D universal features can effectively build correspondences with geometric semantic similarity, thereby constraining the ambiguity caused by optimizing the shape and pose.

\noindent \textbf{Effect of 2D Universal Features.}
We evaluate different combinations of image features for object pose estimation without iteration and pose refinement and the results are shown in Tab.~\ref{table:feature_comb} and Fig.~\ref{fig:ablation}. This experiment shows that DINOv2 is more effective than DINOv1 in capturing semantic similarity, making the predicted scale and rotation more accurate. Furthermore, combining DINOv2 with Stable Diffusion can make semantic correspondences smoother because SD features can take care of global relevance. Note 'ALL' will establish more correspondences on the boundary of object which results in a oversize scale estimation, which results in higher IOU but inaccurate pose.
Therefore, we combine DINOv2 and Stable Diffusion features to establish the most accurate correspondences for coarse pose estimation.

\noindent \textbf{Limitations and Failure Cases.}
Computation time is the bottleneck of our method. In real applications, we can use the full pipeline to locate objects in the first frame and perform efficient pose tracking with few fitting steps in 'Pose Refinement' initialized with pose of previous frame.
The occlusion will also affect the accuracy of the pose estimation results (e.g., as shown in Fig.~\ref{fig:shape_def} (b)). We can use the inpainting method~\cite{suvorov2022resolution} to complete the image to establish correspondences, and then use the visible point cloud for registration.
\section{5. Conclusion}
We propose a new universal features guided zero-shot category-level object pose estimation method in coarse-to-fine fashion. It can estimate the 6D pose of objects from unseen categories without additional model fine-tuning. Our method efficiently utilizes 2D and 3D pre-trained universal features to achieve strong generalization capabilities. It can potentially help many applications deal with unseen categories and avoid additional model training or fine-tuning.

\newpage
\section{Acknowledgments}
This work was supported in part by National Science and Technology Major Project (2022ZD0117904), National Natural Science Foundation of China (62473356,62373061), Beijing Natural Science Foundation (L232028), and CAS Major Project (RCJJ-145-24-14). Heng Li and Ping Tan are supported by the project HKPC22EG01-E from the Hong Kong Industrial Artificial Intelligence \& Robotics Centre (FLAIR).
\bibliography{aaai25}

\begin{thebibliography}{50}
\providecommand{\natexlab}[1]{#1}

\bibitem[{Amir et~al.(2021)Amir, Gandelsman, Bagon, and Dekel}]{amir2021deep}
Amir, S.; Gandelsman, Y.; Bagon, S.; and Dekel, T. 2021.
\newblock Deep vit features as dense visual descriptors.
\newblock \emph{arXiv preprint arXiv:2112.05814}, 2(3): 4.

\bibitem[{Burchfiel and Konidaris(2019)}]{burchfiel2019probabilistic}
Burchfiel, B.; and Konidaris, G. 2019.
\newblock Probabilistic category-level pose estimation via segmentation and predicted-shape priors.
\newblock \emph{arXiv preprint arXiv:1905.12079}.

\bibitem[{Caron et~al.(2021)Caron, Touvron, Misra, J{\'e}gou, Mairal, Bojanowski, and Joulin}]{dinov1}
Caron, M.; Touvron, H.; Misra, I.; J{\'e}gou, H.; Mairal, J.; Bojanowski, P.; and Joulin, A. 2021.
\newblock Emerging properties in self-supervised vision transformers.
\newblock In \emph{Proceedings of the IEEE/CVF International Conference on Computer Vision}, 9650--9660.

\bibitem[{Chen et~al.(2020)Chen, Li, Wang, and Xu}]{chen2020learning}
Chen, D.; Li, J.; Wang, Z.; and Xu, K. 2020.
\newblock Learning canonical shape space for category-level 6d object pose and size estimation.
\newblock In \emph{Proceedings of the IEEE/CVF Conference on Computer Vision and Pattern Recognition}, 11973--11982.

\bibitem[{Chen et~al.(2023)Chen, Sun, Bao, Zhao, Wu, and He}]{chen2023zeropose}
Chen, J.; Sun, M.; Bao, T.; Zhao, R.; Wu, L.; and He, Z. 2023.
\newblock ZeroPose: CAD-model-based zero-shot pose estimation.
\newblock \emph{arXiv preprint arXiv:2305.17934}, 2.

\bibitem[{Chen et~al.(2021)Chen, Jia, Chang, Duan, Shen, and Leonardis}]{fsnet}
Chen, W.; Jia, X.; Chang, H.~J.; Duan, J.; Shen, L.; and Leonardis, A. 2021.
\newblock Fs-net: Fast shape-based network for category-level 6d object pose estimation with decoupled rotation mechanism.
\newblock In \emph{Proceedings of the IEEE/CVF Conference on Computer Vision and Pattern Recognition}, 1581--1590.

\bibitem[{Chen et~al.(2024)Chen, Di, Zhai, Manhardt, Zhang, Zhang, Tombari, Navab, and Busam}]{chen2024secondpose}
Chen, Y.; Di, Y.; Zhai, G.; Manhardt, F.; Zhang, C.; Zhang, R.; Tombari, F.; Navab, N.; and Busam, B. 2024.
\newblock Secondpose: Se (3)-consistent dual-stream feature fusion for category-level pose estimation.
\newblock In \emph{Proceedings of the IEEE/CVF Conference on Computer Vision and Pattern Recognition}, 9959--9969.

\bibitem[{Deitke et~al.(2023)Deitke, Schwenk, Salvador, Weihs, Michel, VanderBilt, Schmidt, Ehsani, Kembhavi, and Farhadi}]{deitke2023objaverse}
Deitke, M.; Schwenk, D.; Salvador, J.; Weihs, L.; Michel, O.; VanderBilt, E.; Schmidt, L.; Ehsani, K.; Kembhavi, A.; and Farhadi, A. 2023.
\newblock Objaverse: A universe of annotated 3d objects.
\newblock In \emph{Proceedings of the IEEE/CVF Conference on Computer Vision and Pattern Recognition}, 13142--13153.

\bibitem[{Di et~al.(2022)Di, Zhang, Lou, Manhardt, Ji, Navab, and Tombari}]{gpv}
Di, Y.; Zhang, R.; Lou, Z.; Manhardt, F.; Ji, X.; Navab, N.; and Tombari, F. 2022.
\newblock Gpv-pose: Category-level object pose estimation via geometry-guided point-wise voting.
\newblock In \emph{Proceedings of the IEEE/CVF Conference on Computer Vision and Pattern Recognition}, 6781--6791.

\bibitem[{Fan et~al.(2023)Fan, Pan, Wang, Jiang, Xu, Jiang, and Wang}]{fan2023pope}
Fan, Z.; Pan, P.; Wang, P.; Jiang, Y.; Xu, D.; Jiang, H.; and Wang, Z. 2023.
\newblock POPE: 6-DoF Promptable Pose Estimation of Any Object, in Any Scene, with One Reference.
\newblock \emph{arXiv preprint arXiv:2305.15727}.

\bibitem[{Fischler and Bolles(1981)}]{fischler1981random}
Fischler, M.~A.; and Bolles, R.~C. 1981.
\newblock Random sample consensus: a paradigm for model fitting with applications to image analysis and automated cartography.
\newblock \emph{Communications of the ACM}, 24(6): 381--395.

\bibitem[{Goodwin et~al.(2022)Goodwin, Vaze, Havoutis, and Posner}]{goodwin2022zero}
Goodwin, W.; Vaze, S.; Havoutis, I.; and Posner, I. 2022.
\newblock Zero-shot category-level object pose estimation.
\newblock In \emph{Proceedings of the European Conference on Computer Vision (ECCV)}, 516--532. Springer.

\bibitem[{He et~al.(2017)He, Gkioxari, Doll{\'a}r, and Girshick}]{maskrcnn}
He, K.; Gkioxari, G.; Doll{\'a}r, P.; and Girshick, R. 2017.
\newblock Mask r-cnn.
\newblock In \emph{Proceedings of the IEEE International Conference on Computer Vision}, 2961--2969.

\bibitem[{He et~al.(2022)He, Fan, Huang, Chen, and Sun}]{he2022towards}
He, Y.; Fan, H.; Huang, H.; Chen, Q.; and Sun, J. 2022.
\newblock Towards self-supervised category-level object pose and size estimation.
\newblock \emph{arXiv preprint arXiv:2203.02884}.

\bibitem[{Hodan, Barath, and Matas(2020)}]{hodan2020epos}
Hodan, T.; Barath, D.; and Matas, J. 2020.
\newblock Epos: Estimating 6d pose of objects with symmetries.
\newblock In \emph{Proceedings of the IEEE/CVF Conference on Computer Vision and Pattern Recognition}, 11703--11712.

\bibitem[{Hu et~al.(2020)Hu, Fua, Wang, and Salzmann}]{hu2020single}
Hu, Y.; Fua, P.; Wang, W.; and Salzmann, M. 2020.
\newblock Single-stage 6d object pose estimation.
\newblock In \emph{Proceedings of the IEEE/CVF Conference on Computer Vision and Pattern Recognition}, 2930--2939.

\bibitem[{Huang, Su, and Guibas(2018)}]{huang2018robust}
Huang, J.; Su, H.; and Guibas, L. 2018.
\newblock Robust watertight manifold surface generation method for shapenet models.
\newblock \emph{arXiv preprint arXiv:1802.01698}.

\bibitem[{Jung et~al.(2024)Jung, Wu, Ruhkamp, Zhai, Schieber, Rizzoli, Wang, Zhao, Garattoni, Meier et~al.}]{jung2024housecat6d}
Jung, H.; Wu, S.-C.; Ruhkamp, P.; Zhai, G.; Schieber, H.; Rizzoli, G.; Wang, P.; Zhao, H.; Garattoni, L.; Meier, S.; et~al. 2024.
\newblock HouseCat6D-A Large-Scale Multi-Modal Category Level 6D Object Perception Dataset with Household Objects in Realistic Scenarios.
\newblock In \emph{Proceedings of the IEEE/CVF Conference on Computer Vision and Pattern Recognition}, 22498--22508.

\bibitem[{Kingma and Ba(2014)}]{adam}
Kingma, D.~P.; and Ba, J. 2014.
\newblock Adam: A method for stochastic optimization.
\newblock \emph{arXiv preprint arXiv:1412.6980}.

\bibitem[{Labb{\'e} et~al.(2020)Labb{\'e}, Carpentier, Aubry, and Sivic}]{cosypose}
Labb{\'e}, Y.; Carpentier, J.; Aubry, M.; and Sivic, J. 2020.
\newblock Cosypose: Consistent multi-view multi-object 6d pose estimation.
\newblock In \emph{Computer Vision--ECCV 2020: 16th European Conference, Glasgow, UK, August 23--28, 2020, Proceedings, Part XVII 16}, 574--591. Springer.

\bibitem[{Labb{\'e} et~al.(2023)Labb{\'e}, Manuelli, Mousavian, Tyree, Birchfield, Tremblay, Carpentier, Aubry, Fox, and Sivic}]{labbe2023megapose}
Labb{\'e}, Y.; Manuelli, L.; Mousavian, A.; Tyree, S.; Birchfield, S.; Tremblay, J.; Carpentier, J.; Aubry, M.; Fox, D.; and Sivic, J. 2023.
\newblock MegaPose: 6D Pose Estimation of Novel Objects via Render \& Compare.
\newblock In \emph{Conference on Robot Learning}, 715--725. PMLR.

\bibitem[{Li et~al.(2018)Li, Wang, Ji, Xiang, and Fox}]{li2018deepim}
Li, Y.; Wang, G.; Ji, X.; Xiang, Y.; and Fox, D. 2018.
\newblock Deepim: Deep iterative matching for 6d pose estimation.
\newblock In \emph{Proceedings of the European Conference on Computer Vision (ECCV)}, 683--698.

\bibitem[{Lin et~al.(2021)Lin, Liu, Cheang, Zhang, Fu, and Xue}]{lin2021donet}
Lin, H.; Liu, Z.; Cheang, C.; Zhang, L.; Fu, Y.; and Xue, X. 2021.
\newblock Donet: Learning category-level 6d object pose and size estimation from depth observation.
\newblock \emph{arXiv preprint arXiv:2106.14193}, 4: 11--12.

\bibitem[{Lin et~al.(2022{\natexlab{a}})Lin, Wei, Ding, and Jia}]{dpdn}
Lin, J.; Wei, Z.; Ding, C.; and Jia, K. 2022{\natexlab{a}}.
\newblock Category-level 6D object pose and size estimation using self-supervised deep prior deformation networks.
\newblock In \emph{European Conference on Computer Vision}, 19--34. Springer.

\bibitem[{Lin et~al.(2023)Lin, Wei, Zhang, and Jia}]{vinet}
Lin, J.; Wei, Z.; Zhang, Y.; and Jia, K. 2023.
\newblock Vi-net: Boosting category-level 6d object pose estimation via learning decoupled rotations on the spherical representations.
\newblock In \emph{Proceedings of the IEEE/CVF International Conference on Computer Vision}, 14001--14011.

\bibitem[{Lin et~al.(2024)Lin, Yang, Gao, and Zhang}]{lin2024instance}
Lin, X.; Yang, W.; Gao, Y.; and Zhang, T. 2024.
\newblock Instance-adaptive and geometric-aware keypoint learning for category-level 6d object pose estimation.
\newblock In \emph{Proceedings of the IEEE/CVF Conference on Computer Vision and Pattern Recognition}, 21040--21049.

\bibitem[{Lin et~al.(2022{\natexlab{b}})Lin, Tremblay, Tyree, Vela, and Birchfield}]{lin2022single}
Lin, Y.; Tremblay, J.; Tyree, S.; Vela, P.~A.; and Birchfield, S. 2022{\natexlab{b}}.
\newblock Single-stage keypoint-based category-level object pose estimation from an RGB image.
\newblock In \emph{2022 International Conference on Robotics and Automation (ICRA)}, 1547--1553. IEEE.

\bibitem[{Luo et~al.(2023)Luo, Dunlap, Park, Holynski, and Darrell}]{luo2023diffusion}
Luo, G.; Dunlap, L.; Park, D.~H.; Holynski, A.; and Darrell, T. 2023.
\newblock Diffusion Hyperfeatures: Searching Through Time and Space for Semantic Correspondence.
\newblock \emph{arXiv preprint arXiv:2305.14334}.

\bibitem[{Oquab et~al.(2023)Oquab, Darcet, Moutakanni, Vo, Szafraniec, Khalidov, Fernandez, Haziza, Massa, El-Nouby et~al.}]{dinov2}
Oquab, M.; Darcet, T.; Moutakanni, T.; Vo, H.; Szafraniec, M.; Khalidov, V.; Fernandez, P.; Haziza, D.; Massa, F.; El-Nouby, A.; et~al. 2023.
\newblock Dinov2: Learning robust visual features without supervision.
\newblock \emph{arXiv preprint arXiv:2304.07193}.

\bibitem[{{\"O}rnek et~al.(2023){\"O}rnek, Labb{\'e}, Tekin, Ma, Keskin, Forster, and Hodan}]{ornek2023foundpose}
{\"O}rnek, E.~P.; Labb{\'e}, Y.; Tekin, B.; Ma, L.; Keskin, C.; Forster, C.; and Hodan, T. 2023.
\newblock FoundPose: Unseen Object Pose Estimation with Foundation Features.
\newblock \emph{arXiv preprint arXiv:2311.18809}.

\bibitem[{Peng et~al.(2019)Peng, Liu, Huang, Zhou, and Bao}]{peng2019pvnet}
Peng, S.; Liu, Y.; Huang, Q.; Zhou, X.; and Bao, H. 2019.
\newblock Pvnet: Pixel-wise voting network for 6dof pose estimation.
\newblock In \emph{Proceedings of the IEEE/CVF Conference on Computer Vision and Pattern Recognition}, 4561--4570.

\bibitem[{Peng et~al.(2022)Peng, Yan, Wen, and Sun}]{SSC}
Peng, W.; Yan, J.; Wen, H.; and Sun, Y. 2022.
\newblock Self-supervised category-level 6D object pose estimation with deep implicit shape representation.
\newblock In \emph{Proceedings of the AAAI Conference on Artificial Intelligence}, volume~36, 2082--2090.

\bibitem[{Ravi et~al.(2020)Ravi, Reizenstein, Novotny, Gordon, Lo, Johnson, and Gkioxari}]{ravi2020pytorch3d}
Ravi, N.; Reizenstein, J.; Novotny, D.; Gordon, T.; Lo, W.-Y.; Johnson, J.; and Gkioxari, G. 2020.
\newblock Accelerating 3D Deep Learning with PyTorch3D.
\newblock \emph{arXiv:2007.08501}.

\bibitem[{Rombach et~al.(2022)Rombach, Blattmann, Lorenz, Esser, and Ommer}]{stable_diffusion}
Rombach, R.; Blattmann, A.; Lorenz, D.; Esser, P.; and Ommer, B. 2022.
\newblock High-resolution image synthesis with latent diffusion models.
\newblock In \emph{Proceedings of the IEEE/CVF Conference on Computer Vision and Pattern Recognition}, 10684--10695.

\bibitem[{Sahin and Kim(2018)}]{sahin2018category}
Sahin, C.; and Kim, T.-K. 2018.
\newblock Category-level 6d object pose recovery in depth images.
\newblock In \emph{Proceedings of the European Conference on Computer Vision (ECCV) Workshops}, 0--0.

\bibitem[{Suvorov et~al.(2022)Suvorov, Logacheva, Mashikhin, Remizova, Ashukha, Silvestrov, Kong, Goka, Park, and Lempitsky}]{suvorov2022resolution}
Suvorov, R.; Logacheva, E.; Mashikhin, A.; Remizova, A.; Ashukha, A.; Silvestrov, A.; Kong, N.; Goka, H.; Park, K.; and Lempitsky, V. 2022.
\newblock Resolution-robust large mask inpainting with fourier convolutions.
\newblock In \emph{Proceedings of the IEEE/CVF winter conference on applications of computer vision}, 2149--2159.

\bibitem[{Tang et~al.(2023)Tang, Jia, Wang, Phoo, and Hariharan}]{tang2023emergent}
Tang, L.; Jia, M.; Wang, Q.; Phoo, C.~P.; and Hariharan, B. 2023.
\newblock Emergent Correspondence from Image Diffusion.
\newblock \emph{arXiv preprint arXiv:2306.03881}.

\bibitem[{Tekin, Sinha, and Fua(2018)}]{tekin2018real}
Tekin, B.; Sinha, S.~N.; and Fua, P. 2018.
\newblock Real-time seamless single shot 6d object pose prediction.
\newblock In \emph{Proceedings of the IEEE Conference on Computer Vision and Pattern Recognition}, 292--301.

\bibitem[{Tian, Ang, and Lee(2020)}]{tian2020shape}
Tian, M.; Ang, M.~H.; and Lee, G.~H. 2020.
\newblock Shape prior deformation for categorical 6d object pose and size estimation.
\newblock In \emph{Computer Vision--ECCV 2020: 16th European Conference, Glasgow, UK, August 23--28, 2020, Proceedings, Part XXI 16}, 530--546. Springer.

\bibitem[{Umeyama(1991)}]{umeyama1991least}
Umeyama, S. 1991.
\newblock Least-squares estimation of transformation parameters between two point patterns.
\newblock \emph{IEEE Transactions on Pattern Analysis \& Machine Intelligence}, 13(04): 376--380.

\bibitem[{Wang et~al.(2021)Wang, Manhardt, Tombari, and Ji}]{wang2021gdr}
Wang, G.; Manhardt, F.; Tombari, F.; and Ji, X. 2021.
\newblock Gdr-net: Geometry-guided direct regression network for monocular 6d object pose estimation.
\newblock In \emph{Proceedings of the IEEE/CVF Conference on Computer Vision and Pattern Recognition}, 16611--16621.

\bibitem[{Wang et~al.(2019{\natexlab{a}})Wang, Sridhar, Huang, Valentin, Song, and Guibas}]{normalized}
Wang, H.; Sridhar, S.; Huang, J.; Valentin, J.; Song, S.; and Guibas, L.~J. 2019{\natexlab{a}}.
\newblock Normalized object coordinate space for category-level 6d object pose and size estimation.
\newblock In \emph{Proceedings of the IEEE/CVF Conference on Computer Vision and Pattern Recognition}, 2642--2651.

\bibitem[{Wang et~al.(2019{\natexlab{b}})Wang, Sridhar, Huang, Valentin, Song, and Guibas}]{wang2019normalized}
Wang, H.; Sridhar, S.; Huang, J.; Valentin, J.; Song, S.; and Guibas, L.~J. 2019{\natexlab{b}}.
\newblock Normalized object coordinate space for category-level 6d object pose and size estimation.
\newblock In \emph{Proceedings of the IEEE/CVF Conference on Computer Vision and Pattern Recognition}, 2642--2651.

\bibitem[{Wang et~al.(2022)Wang, Jung, Li, Shen, Srikanth, Garattoni, Meier, Navab, and Busam}]{wang2022phocal}
Wang, P.; Jung, H.; Li, Y.; Shen, S.; Srikanth, R.~P.; Garattoni, L.; Meier, S.; Navab, N.; and Busam, B. 2022.
\newblock Phocal: A multi-modal dataset for category-level object pose estimation with photometrically challenging objects.
\newblock In \emph{Proceedings of the IEEE/CVF conference on computer vision and pattern recognition}, 21222--21231.

\bibitem[{Wang et~al.(2019{\natexlab{c}})Wang, Sun, Liu, Sarma, Bronstein, and Solomon}]{dgcnn}
Wang, Y.; Sun, Y.; Liu, Z.; Sarma, S.~E.; Bronstein, M.~M.; and Solomon, J.~M. 2019{\natexlab{c}}.
\newblock Dynamic graph cnn for learning on point clouds.
\newblock \emph{ACM Transactions on Graphics (tog)}, 38(5): 1--12.

\bibitem[{Wen et~al.(2024)Wen, Yang, Kautz, and Birchfield}]{wen2024foundationpose}
Wen, B.; Yang, W.; Kautz, J.; and Birchfield, S. 2024.
\newblock Foundationpose: Unified 6d pose estimation and tracking of novel objects.
\newblock In \emph{Proceedings of the IEEE/CVF Conference on Computer Vision and Pattern Recognition}, 17868--17879.

\bibitem[{Xiang et~al.(2017)Xiang, Schmidt, Narayanan, and Fox}]{posecnn}
Xiang, Y.; Schmidt, T.; Narayanan, V.; and Fox, D. 2017.
\newblock Posecnn: A convolutional neural network for 6d object pose estimation in cluttered scenes.
\newblock \emph{arXiv preprint arXiv:1711.00199}.

\bibitem[{Ze and Wang(2022)}]{ze2022category}
Ze, Y.; and Wang, X. 2022.
\newblock Category-level 6d object pose estimation in the wild: A semi-supervised learning approach and a new dataset.
\newblock \emph{Advances in Neural Information Processing Systems}, 35: 27469--27483.

\bibitem[{Zhang et~al.(2023)Zhang, Herrmann, Hur, Cabrera, Jampani, Sun, and Yang}]{zhang2023tale}
Zhang, J.; Herrmann, C.; Hur, J.; Cabrera, L.~P.; Jampani, V.; Sun, D.; and Yang, M.-H. 2023.
\newblock A Tale of Two Features: Stable Diffusion Complements DINO for Zero-Shot Semantic Correspondence.
\newblock \emph{arXiv preprint arXiv:2305.15347}.

\bibitem[{Zhang et~al.(2022)Zhang, Fu, Borse, Cai, Porikli, and Wang}]{zhang2022self}
Zhang, K.; Fu, Y.; Borse, S.; Cai, H.; Porikli, F.; and Wang, X. 2022.
\newblock Self-supervised geometric correspondence for category-level 6d object pose estimation in the wild.
\newblock \emph{arXiv preprint arXiv:2210.07199}.

\end{thebibliography}
\clearpage
\appendix
\section{Appendix}
In this supplementary material, we first introduce the implementation details of our method (Sec.~1). Then we show additional comparison experimental details (Sec.~2) and ablation experimental results (Sec.~3).

\section{1. Method Details}
\label{sec:supp_method}

\subsection{1.1 Initial View of Reference Images}
\begin{figure*}[ht]
\centering%
\includegraphics[width=\linewidth]{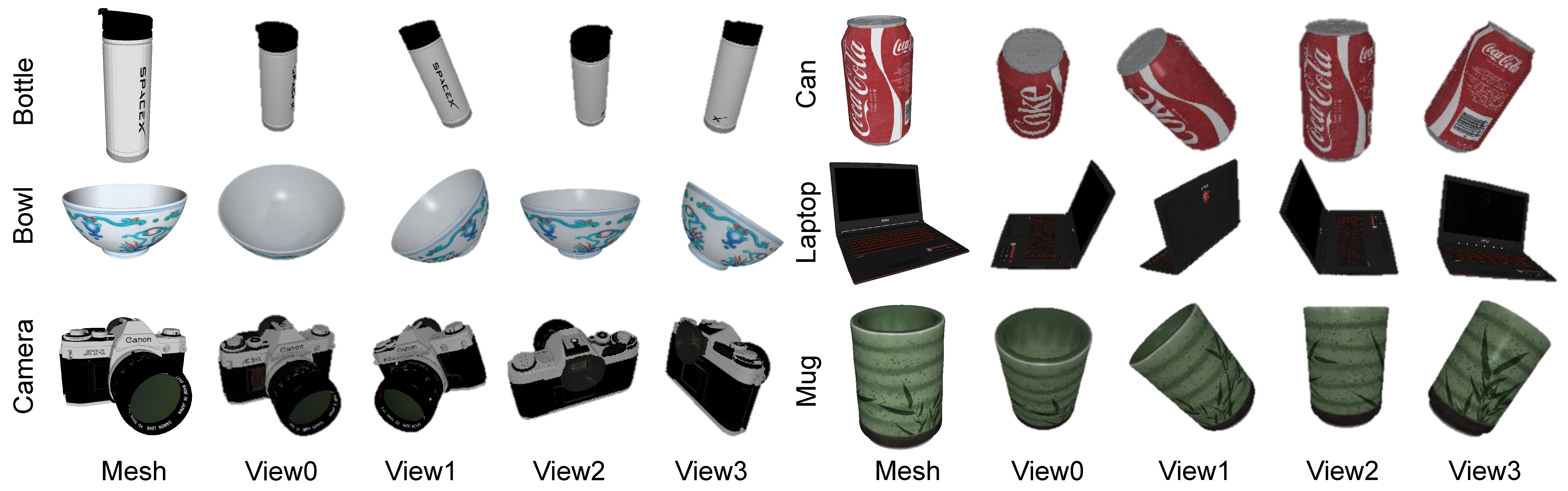}
\caption{We render four reference images of each reference 3D model from the front, back, and two sides, as input to 'Coarse Pose Estimation' stage.}
\label{fig:supp_ref_views}
\end{figure*}

We render four reference images $\{ \mathbf{I}_{r} \}$ of the reference 3D model from the front, back, and two sides. 
We show the four views of the six category objects in Fig.~\ref{fig:supp_ref_views}, respectively. 
They have great differences in texture and geometry from the instance objects in the benchmarks.

\subsection{1.2 Details on Multi-Modal Universal Features}

The reference mesh model can be retrieved from the Objverse~\cite{deitke2023objaverse}, and we use Pytorch3d~\cite{ravi2020pytorch3d} to render the reference RGB-D images. 
For 2D universal feature extraction, we follow the methods~\cite{goodwin2022zero,zhang2023tale,luo2023diffusion} and use DINOv1(`vit\_small'), DINOv2(`vits14') and SD(`v1-5').
The 3D universal features are extracted from \emph{conv6} of the pre-trained DGCNN~\cite{dgcnn}.

The dimensions of the 2D universal feature $F_{D1}$, $F_{D2}$ and $F_{SD}$ are 6528, 384 and 10560, respectively, and for each 2D universal feature pair of the reference image and target image, we reduce its dimension to 64 using PCA. We also reduce the dimensions of 3D universal features from 1024 to 64. 

Since several meshes from Objaverse~\cite{deitke2023objaverse} are not watertight, we use the method~\cite{huang2018robust} to process these meshes into watertight meshes before feeding them into the pose refinement stage.

\subsection{1.3 Details on Regularization Loss $L_{r}$.}
The regularization loss includes pose regularization loss $L_{p}$, center point regularization loss $L_{ce}$, and deformation regularization loss $L_{d}$. We also follow Pytorch3D~\cite{ravi2020pytorch3d} to minimize geometric distortion with normal, edge, and Laplacian constraints.
Our regularization loss is defined as: $L_{r} = \alpha_{p}L_{p} + \alpha_{ce}L_{ce} + \alpha_{d}L_{d}$.

The pose regularization loss $L_p$ enforces the refined pose to be close to the initial pose computed by the coarse estimation module, balancing the contributions between image and point cloud features: $L_{p} = \Vert (\Bar{\mathbf{R}} \Bar{\mathbf{V}}+\Bar{\mathbf{T}}) - (\hat{\mathbf{R}}\mathbf{V}+\hat{\mathbf{T}}) \Vert_{2}$.

The center point regularization loss $L_{ce}$ encourages the displacement of object before and after optimization to be small as: $L_{ce} = \Vert C(\Bar{\mathbf{R}}\Bar{\mathbf{V}}+\Bar{\mathbf{T}})- C(\hat{\mathbf{R}}\mathbf{V}+\hat{\mathbf{T}})  \Vert_{2}$, 
where $C(\cdot)$ means average.

The deformation regularization loss $L_{d}=\Vert \Delta \mathbf{V} \Vert_{2}$ is used to constrain the deformation of vertices to be small.

We set the hyperparameters 
$\alpha_{p}$,  $\alpha_{ce}$, $\alpha_{d}$ to 20, 1 and 1 respectively.

\begin{table*}[h]
\centering

\fontsize{9}{11}\selectfont
\begin{tabular}{c|c|c|c|c|c|c|c}
\hline
{Dataset}& {Model Type}&$IOU_{0.25}\uparrow$  & $IOU_{0.5}\uparrow$ & $5^{\circ}*2cm\uparrow$ & $5^{\circ}*5cm\uparrow$ & $10^{\circ}*2cm\uparrow$ & $10^{\circ}*5cm\uparrow$\\
\hline
\multirow{2}{*}{{\begin{tabular}[c]{@{}c@{}}REAL275\end{tabular}}}&
{Supplementary}&{75.08}&\textbf{63.49}&{29.41}&\textbf{34.37}&{44.74}&\textbf{59.17}\\
{}&{Main Paper}&\textbf{80.06}&\textbf{63.49}&\textbf{30.61}&{33.23}&\textbf{50.15}&{57.74}\\
\hline\hline
\multirow{2}{*}{{\begin{tabular}[c]{@{}c@{}}Wild6D\end{tabular}}}&
{Supplementary}&{85.38}&\textbf{70.83}&{46.57}&{57.49}&\textbf{60.57}&{78.59}\\
{}&{Main Paper}&\textbf{88.46}&{67.16}&\textbf{47.69}&\textbf{60.58}&{59.46}&\textbf{80.47}\\
\hline
\end{tabular}
\caption{Effect of reference 3D model. We change the model from Main Paper to Supplementary and find that our method is not sensitive to a specific model and shows strong generalization ability.}
\label{table:supp_generic_model}
\end{table*}

\section{2. Comparison Experiment Details}
\label{sec:supp_com}

\subsection{2.1 Baseline Implementation Details} 
\label{sec:supp_exp_implement_details}
We compare our method with three types of baselines: 1) Supervised methods: DPDN~\cite{dpdn}, VI-Net~\cite{vinet} and SPD~\cite{tian2020shape}. 2) Self-Supervised methods: Self-Pose~\cite{zhang2022self}, Wild6D~\cite{ze2022category} ans SSC-6D~\cite{SSC}. 3) Zero-Shot method: Zero-Pose~\cite{goodwin2022zero} and MegaPose~\cite{labbe2023megapose}. 
We elaborate on the training and testing settings for the compared methods in Tab.1 of the main paper. We use the test sets of REAL275~\cite{wang2019normalized} and Wild6D~\cite{ze2022category} for testing.

For the supervised methods, we adapt the leave-1 strategy, which is to select one category as the test set and use the remaining categories to train the model. This ensures that the model can be tested on unseen categories. We conduct leave-1 experiments for each category and finally take the average as the final results.
We train DPDN~\cite{dpdn}, VI-Net~\cite{vinet} and SPD~\cite{tian2020shape} on synthetic CAMERA25 and real-world REAL275 datasets~\cite{wang2019normalized}. 

For the self-supervised methods Self-Pose~\cite{zhang2022self}, Wild6D~\cite{ze2022category} ans SSC-6D~\cite{SSC}, we directly use the officially trained model to conduct the leave-p experiments. Because their solution requires prior information of category objects, and each category object model needs to be trained separately. 
For the category to be tested, we perform other category models on it and take the average as the final results.

For the Zero-shot method, Zero-Pose uses the same reference model and rendered reference image as the initial input as our method, and MegaPose uses the same reference model as our method.

In order to verify the generalization ability of our method, we choose a reference mug model without handle, resulting in symmetrical properties. 
We also find that there is a lack of handle visibility annotation in the mug of Wild6D benchmark~\cite{ze2022category}, which will lead to pose accuracy deviation if we consider all mugs as asymmetric objects (e.g., a mug without a handle is a symmetric object).
For a fair comparison, we treat mugs as symmetrical objects on all benchmarks, which will not affect the performance of comparison methods.

\subsection{2.2 Annotations in WILD6D dataset} 
We find that some annotations in the Wild6D~\cite{ze2022category} test set are not aligned to the NOCS coordinate~\cite{wang2019normalized}, which is quite different from the annotations of REAL275~\cite{wang2019normalized}. So we perform low-cost processing on it and unify the annotations into the NOCS coordinate.

\subsection{2.3 Additional Comparison Experiment}
To verify the effectiveness of our method, we test novel unseen categories 'teapot' from 'test\_scene4' and 'tube' from 'test\_scene3' on HouseCat6D dataset~\cite{jung2024housecat6d}. We compare with DPDN~\cite{dpdn}, VI-Net~\cite{vinet}, Wild6D~\cite{ze2022category}, SSC-6D~\cite{SSC} and Zero-Pose~\cite{goodwin2022zero}, and the results are shown in Table~\ref{table:supp_housecat6d}.

\begin{table*}[h]
\centering

\fontsize{9}{11}\selectfont
\begin{tabular}{c|c|c|c|c|c|c|c}
\hline
{Dataset}& {Method}&$IOU_{0.25}\uparrow$  & $IOU_{0.5}\uparrow$ & $5^{\circ}*2cm\uparrow$ & $5^{\circ}*5cm\uparrow$ & $10^{\circ}*2cm\uparrow$ & $10^{\circ}*5cm\uparrow$\\
\hline
\multirow{6}{*}{{\begin{tabular}[c]{@{}c@{}}HouseCat6D\end{tabular}}}&
   {DPDN}&{80.94}&{3.37}&{2.75}&{8.45}&{8.53}&{20.76}\\
{}&{VI-Net}&{80.54}&{12.33}&{13.62}&{43.45}&{15.73}&{52.54}\\
{}&{Wild6D}&{56.14}&{13.64}&{1.92}&{7.30}&{6.70}&{22.24}\\
{}&{SSC-6D}&{61.59}&{3.59}&{1.08}&{3.87}&{4.12}&{14.64}\\
{}&{Zero-Pose}&{97.87}&{48.04}&{6.46}&{11.39}&{35.03}&{59.10}\\
{}&{Ours}&\textbf{99.15}&\textbf{85.37}&\textbf{43.20}&\textbf{46.00}&\textbf{69.30}&\textbf{76.11}\\
\hline
\end{tabular}
\caption{Comparison results on HouseCat6D. Our method outperforms other methods on novel unseen categories, verifying the robustness of our method to unseen categories.}
\label{table:supp_housecat6d}
\end{table*}

For the comparison methods, we use the official pre-trained model for testing. We find that our method can still perform better when facing unseen categories because our method achieves pose estimation by finding the correspondence between objects.

\section{3. Additional Ablation Experiments}
\label{sec:supp_abl}

\subsection{3.1 Effect of Reference 3D Model}
\begin{figure}[ht]
\centering%
\includegraphics[width=\linewidth]{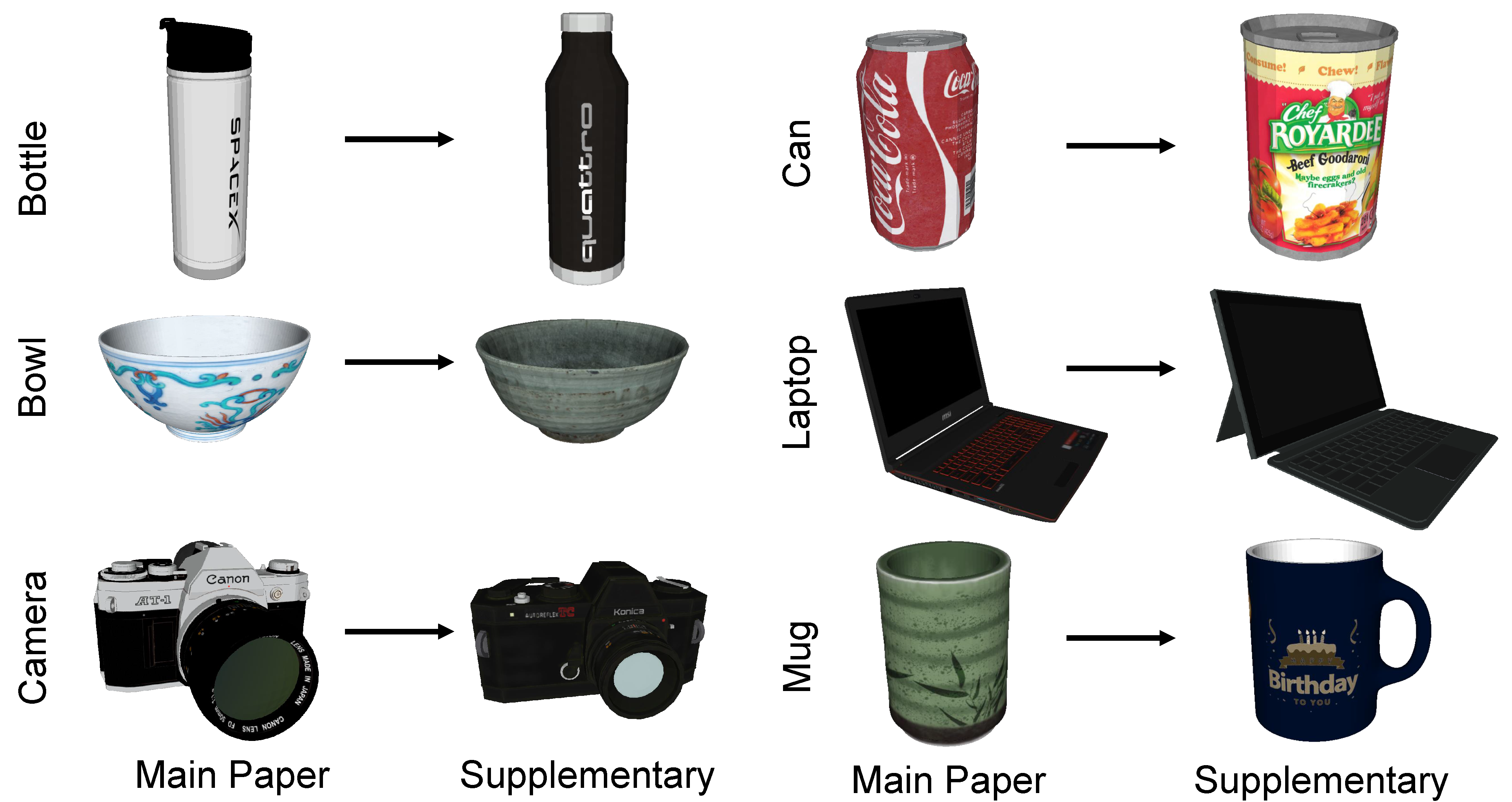}
\caption{We change the reference models from used in main paper to used in supplementary with significantly different texture and shape, to verify that our method is not restricted by a specially defined model.}
\label{fig:supp_model_change}
\end{figure}

To demonstrate that our pipeline is not sensitive to the specific geometry and texture of the reference 3D model, we replace the objects used in the main paper with significantly different objects as shown in Fig.~\ref{fig:supp_model_change}, and we show the results in Tab.~\ref{table:supp_generic_model}. Our method still shows the same effect when replaced with a new reference model. Therefore, our method is not restricted by a specific reference object model and has superior generalization ability.

\begin{table*}[h]
\centering
\fontsize{9}{11}\selectfont
\begin{tabular}{c|c|c|c|c|c|c|c|c}
\hline
{Dataset}& {Steps}&Speed&$IOU_{0.25}$  & $IOU_{0.5}$ & $5^{\circ}*2cm$ & $5^{\circ}*5cm$ & $10^{\circ}*2cm$ & $10^{\circ}*5cm$\\
\hline
\multirow{6}{*}{\rotatebox{90}{\begin{tabular}[c]{@{}c@{}}REAL275\end{tabular}}}&
{0}&{-}&{80.00}&{58.39}&{28.56}&{30.94}&{48.33}&{55.27}\\
{}&{35}&{0.6s}&{79.99}&{61.32}&{30.00}&{32.74}&{49.25}&{57.39}\\
{}&{50}&{0.9s}&{80.05}&{62.40}&{30.54}&{33.17}&{49.71}&{57.60}\\
{}&{80}&{1.4s}&{80.06}&{63.49}&{30.61}&{33.23}&{50.15}&\textbf{57.74}\\
{}&{110}&{1.9s}&\textbf{80.07}&{63.96}&{30.96}&{33.61}&{50.17}&{57.51}\\
{}&{140}&{2.4s}&{80.02}&\textbf{64.06}&\textbf{31.03}&\textbf{33.69}&\textbf{50.38}&\textbf{57.74}\\
\hline\hline
\multirow{6}{*}{\rotatebox{90}{\begin{tabular}[c]{@{}c@{}}Wild6D\end{tabular}}}&
{0}&{-}&{88.17}&{62.77}&{42.71}&{57.58}&{55.99}&{80.69}\\
{}&{35}&{0.6s}&{88.37}&{65.60}&{45.86}&{59.83}&{58.55}&\textbf{80.79}\\
{}&{50}&{0.9s}&{88.41}&{66.17}&{46.46}&{60.12}&{58.87}&{80.67}\\
{}&{80}&{1.4s}&{88.46}&{67.16}&{47.69}&{60.58}&{59.46}&{80.47}\\
{}&{110}&{1.9s}&\textbf{88.51}&{67.93}&{48.61}&\textbf{60.78}&{60.18}&{80.34}\\
{}&{140}&{2.4s}&{88.50}&\textbf{68.83}&\textbf{49.08}&{60.44}&\textbf{61.05}&{80.19}\\
\hline
\end{tabular}
\caption{Effect of fitting steps in 'Pose Refinement' stage. The accuracy of pose increases as the number of fitting steps increases. In order to keep a good balance between time cost and pose accuracy, we choose 80 fitting steps in our 'Pose Refinement' stage. In the process of pose tracking, we also can choose fewer fitting steps to improve efficiency.}
\label{table:supp_fit_step}
\end{table*}

\subsection{3.2 Effect of Fitting Steps in Pose Refinement}
To investigate the impact of different fitting steps in the pose refinement stage, we conduct experiments under different fitting steps and the results are shown in Tab.~\ref{table:supp_fit_step}.
The accuracy of pose increases as the number of fitting steps increases. In order to keep a good balance between time cost and pose accuracy, we choose 80 fitting steps in our pose refinement stage.

\begin{figure}[ht]
\centering%
\includegraphics[width=\linewidth]{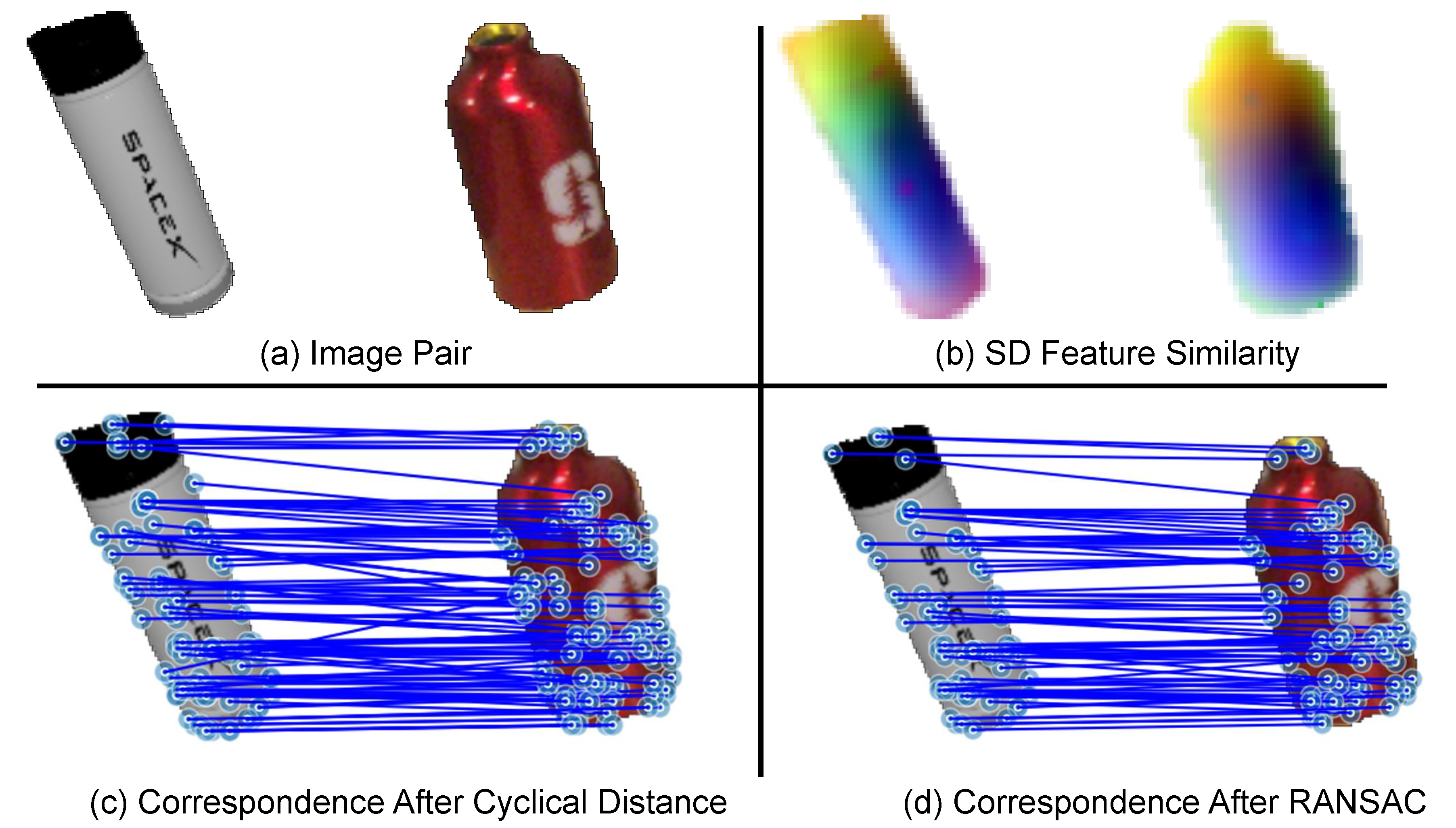}
\caption{Discussion of symmetric objects. The SD universal features can capture global correlation of objects and support the establishment of accurate correspondences. At the same time, RANSAC can be used to eliminate outliers caused by symmetry.}
\label{fig:sym_obj}
\end{figure}

\subsection{3.3 Discussion of Symmetric Objects}
For the task of finding correspondences between different symmetric objects (Fig.~\ref{fig:sym_obj} (a)), we observe that the SD universal features can capture global correlations (Fig.~\ref{fig:sym_obj} (b)) and can be combined with DINOv2 universal features to establish accurate correspondences under low cyclical distance (Fig.~\ref{fig:sym_obj} (c)). In the coarse pose estimation stage,  RANSAC~\cite{fischler1981random} can help us further eliminate outliers due to symmetry (Fig.~\ref{fig:sym_obj} (d)).
Therefore, we can maintain the accuracy of pose estimation for symmetric objects.

\end{document}